\documentstyle[jair,twoside,11pt,theapa]{article}

\jairheading{5}{1996}{53-94}{3/95}{9/96}
\ShortHeadings{Cue Phrase Classification Using Machine Learning}
{Litman}
\firstpageno{53}

\begin{document}

\title{Cue Phrase Classification Using Machine Learning}

\author{\name Diane J. Litman \email diane@research.att.com\\
	\addr AT\&T Labs - Research, 600 Mountain Avenue\\
	Murray Hill, NJ  07974 USA}

\maketitle

\begin{abstract}
Cue phrases may be used in a {\it discourse} sense to explicitly
signal discourse structure, but also in a {\it sentential} sense to
convey semantic rather than structural information.  
Correctly classifying cue phrases as discourse or
sentential is critical in natural language processing systems that exploit discourse structure,
e.g., for performing tasks such as anaphora resolution and plan recognition.
This  paper explores the
use of machine learning for classifying cue phrases as
discourse or sentential.  Two machine learning programs ({\sc
cgrendel} and C4.5) are used to induce classification models from 
sets of pre-classified cue phrases and their features in text and speech.
Machine learning is shown to be an
effective technique for not only automating the generation of
classification models, but also for improving upon previous results.
When compared to manually derived classification models already in the literature, the
learned models often perform with higher accuracy and contain new linguistic insights into
the data.
In addition, the
ability to automatically construct classification models makes it
easier to comparatively analyze the utility of alternative feature
representations of the data.  Finally, the ease of retraining makes
the learning approach more scalable and flexible than manual
methods.
\end{abstract}

\section{Introduction}
\label{introduction}

{\it Cue phrases} are words and phrases that may sometimes be
used to explicitly signal discourse structure in both text and speech.  In particular,
when used in a {\it discourse} sense, a cue phrase explicitly
conveys structural information. When used in a {\it sentential} sense, a cue
phrase instead conveys semantic rather than structural information.  
The following examples (taken from a spoken language corpus that will be described 
in Section~\ref{data}) illustrate sample discourse and sentential usages of the cue phrases
``say'' and ``further'':
\begin{itemize}
\item
Discourse
\begin{description}
\item[]
``\ldots we might have the concept of {\it say} a researcher who has worked for fifteen years on a certain project \ldots''
\item[]
``{\it Further}, and this is crucial in AI and probably for expert databases as well \ldots''
\end{description}
\item
Sentential
\begin{description}
\item[]
``\ldots let me just {\it say} that it bears a strong resemblance to much of the work that's done in semantic nets and even frames.''
\item[]
``\ldots from a place that is even stranger and {\it further} away \ldots''
\end{description}
\end{itemize}
For example, when used
in the discourse sense, the cue phrase ``say'' conveys 
the structural information that an example is beginning. When 
used in the sentential sense, ``say'' does not convey any structural information and instead functions 
as a verb.  

The ability to correctly classify cue phrases as discourse or
sentential is critical for natural language processing systems that
need to recognize or convey discourse structure, for tasks such as
improving anaphora resolution~\cite{Grosz86Attention,Reichman85}.
Consider the following example, again taken from the corpus that will
be described in Section~\ref{data}\footnote{This example is also described
in more detail by \shortciteA{hl93}.}:
\begin{quote}
If \underline{the system} attempts to hold rules, {\it say} as
\underline{an expert database} for \underline{an expert system}, {\it
then\/} we expect \underline{it} not only to hold the rules but to in
fact apply them for us in appropriate situations.
\end{quote}
In this example, the cue phrases ``say'' and ``then'' are
discourse usages, and explicitly signal the boundaries of an
intervening subtopic in the discourse structure.  Furthermore, the referents of the noun phrases
``the system,'' ``an expert database,'' and ``an expert
system'' are all possible referents for the pronoun ``it.''  With
the structural information conveyed by the cue phrases, the system can
determine that ``the system'' is more relevant for interpreting the
pronoun ``it,'' as both ``an expert database'' and ``an expert
system'' occur within the embedded (and now concluded) subtopic.  Without the cue phrases,
the reasoning required to determine that the referent of the ``the
system'' is the intended referent of ``it'' would be much more complex.

Correctly classifying cue phrases as discourse or sentential is
important for other natural language processing tasks as well. 
The discourse/sentential distinction can be used to improve
the naturalness of synthetic speech in text-to-speech
systems~\cite{hirschberg90}.  Text-to-speech systems generate
synthesized speech from unrestricted text.  If a cue phrase can be
classified as discourse or sentential using features of the input
text, it can then be synthesized using different intonational models
for the discourse and sentential usages.  In addition, by explicitly
identifying rhetorical and other relationships, discourse usages of
cue phrases can be used to improve the coherence of multisentential
texts in natural language generation
systems~\cite{Zukerman86,Moser95}.  Cue phrases can also be
used to reduce the complexity of discourse processing in such areas as
argument understanding~\cite{Cohen84} and plan
recognition~\cite{Litman87Plan,Grosz86Attention}.

While the problem of cue phrase classification has often been
noted~\cite{Grosz86Attention}, until recently, models for classifying
cue phrases were neither developed nor evaluated based on careful
empirical analyses.  Even though the literature
suggests that some features might be useful for cue phrase
classification, there are no quantitative analyses of any
actual classification algorithms that use such features (nor any suggestions
as to how different types of features might be combined).
Most systems that recognize or generate cue
phrases simply assume that discourse uses are utterance or clause
initial~\cite{Reichman85,Zukerman86}.  While there are
empirical studies showing that the intonational prominence of certain
word classes varies with respect to discourse
function~\cite{Halliday76,Altenberg87}, these studies do not
investigate cue phrases per se.  

To address these limitations, \shortciteA{hl93}
conducted several empirical studies specifically addressing 
cue phrase classification in text and speech.  Hirschberg and
Litman pre-classified a set of naturally occurring cue phrases,
described each cue phrase in terms of prosodic and textual features
(the features were posited in the literature or easy to
automatically code), then manually examined the data to construct
classification models that best predicted the classifications from
the feature values.

This paper examines the utility of machine learning for automating the
construction of models for classifying cue phrases from such empirical
data.  A set of experiments are described that use two machine
learning programs, {\sc cgrendel}~\cite{CohenIMLC92,CohenIJCAI93} and
C4.5~\cite{Quinlan93}, to induce classification models from sets of
pre-classified cue phrases and their features.  The features, classes and
training examples used in the studies of \shortciteA{hl93}, as well as 
additional features, classes and training examples, are given as input to
the machine learning programs.  The results are evaluated both
quantitatively and qualitatively, by comparing both the error rates
and the content of the manually derived and learned classification
models.  The experimental results show that machine learning is indeed
an effective technique, not only for automating the generation of
classification models, but also for improving upon previous results.
The accuracy of the
learned classification models is often higher than the accuracy of 
the manually derived models, and the learned models often contain new linguistic
implications. The learning paradigm also makes it easier to compare
the utility of different knowledge sources, and to update the model
given new features, classes, or training data.

The next section summarizes previous work on cue phrase
classification.  Section~\ref{ml} then describes the machine learning
approach to cue phrase classification that is taken in this paper. In
particular, the section describes four sets of experiments that use
machine learning to automatically induce cue phrase classification
models. The types of inputs and outputs of the machine learning programs are
presented, as are the methodologies that are used to
evaluate the results.  Section~\ref{results} presents and discusses
the experimental results, and highlights the many benefits of the
machine learning approach.  Section~\ref{utility} discusses the practical utility
of the results of this paper.
Finally, Section~\ref{related} discusses
the use of machine learning in other studies of discourse, while
Section~\ref{conclusion} concludes.
\section{Previous Work on Classifying Cue Phrases}
\label{data}

This section summarizes Hirschberg's and Litman's empirical studies of the classification of
cue phrases in speech and text~\cite{Hirschberg87Now,hl93,Litman90}.
Hirschberg's and Litman's data (cue phrases taken from corpora of recorded and
transcribed speech, classified as {\it discourse} or {\it sentential}, and coded using both speech-based and text-based features)
will be used to create the input for the machine learning experiments.
Hirschberg's and Litman's results (performance figures for manually developed cue phrase classification 
models) will be used as a benchmark for evaluating the performance of the classification models produced by machine learning.

The first study by
Hirschberg and Litman investigated usage of the cue phrase ``now'' by
multiple speakers in a radio call-in
show~\cite{Hirschberg87Now}.  A classification model based on
prosodic features was developed based on manual analysis of a ``training'' set of 48 examples of ``now'',
then evaluated on a previously unseen test set of 52 examples of ``now''. 
In a follow-up study~\cite{hl93}, Hirschberg and
Litman tested this classification model on a larger set of cue phrases, namely
all single word cue phrases in a technical keynote address by a single speaker. 
This corpus yielded 953 instances of 34 different single word cue phrases
derived from the literature.\footnote{Figure~\ref{features} contains a list of the 34 cue phrases.
\shortciteA{hl93} provide full details regarding the distribution of these cue phrases. The most frequent cue phrase is ``and'',  which occurs 320 times. The next
most frequent cue phrase is ``now'', which occurs 69 times.
``But,''  ``like,'' ``or'' and ``so'' also each occur more than fifty times.  
The four least frequent cue phrases -- ``essentially,'' ``otherwise,'' ``since'' and ``therefore'' -- each occur 2 times.}  Hirschberg and Litman also used
the cue phrases in the first 17 minutes of this corpus 
to develop a complementary cue
phrase classification model based on textual
features~\cite{Litman90}, which they then
tested on the full corpus~\cite{hl93}.
The first
study will be referred to as the ``now'' study, and the follow-up study as the ``multiple cue phrase'' study.
Note that the term ``multiple'' means that 34 different single word cue phrases (as opposed to just the cue phrase
``now'') are considered, not that cue phrases consisting of multiple words (e.g. ``by the way'') are considered.

The method that Hirschberg and Litman used to develop their
prosodic and textual classification models was as follows.  They first
separately classified each example cue phrase in the data as {\it discourse}, {\it sentential} or
{\it ambiguous} while listening to a recording and reading a
transcription.\footnote{The class {\it ambiguous} was not introduced
until the multiple cue phrase study~\cite{hl93,Litman90}.}  
Each example was also described as a set of
prosodic and textual features.\footnote{Although a limited set of textual
features were noted in the ``now'' data, the analysis of the ``now'' data did not yield a textual classification model.}
Previous observations in the literature correlating discourse
structure with prosodic information, and discourse usages of cue
phrases with initial position in a clause, contributed to the choice
of features. 
The set of classified and described examples was then examined in order to manually develop
the classification models shown in Figure~\ref{prosody}.  These models are shown here using decision
trees for ease of comparison with the results of C4.5 and will be explained below.
\begin{figure}[tb]
{\scriptsize
\begin{center}
\begin{tabbing}
ssss \= ssss \= sssssssssssssssssssssssssssssssssssssssssssssssssssssssssssssssssssssssss\= \kill
\underline{Prosodic Model}:\\ \\
{\bf if} composition of intermediate phrase $=$ alone {\bf then}  {\it discourse} \>\>\>(1) \\
{\bf elseif}  composition of intermediate phrase $=$ $\neg$alone {\bf then} \>\>\>(2) \\  
\>{\bf if} position in intermediate phrase $=$ first {\bf then}\>\>(3)\\
\>	\>{\bf if} accent $=$  deaccented {\bf then}  {\it discourse} \>(4) \\ 
\>	\>{\bf elseif} accent $=$   L* {\bf then}  {\it discourse}\>(5) \\
\>	\>{\bf elseif} accent $=$   H* {\bf then}  {\it sentential}\>(6) \\
\>	\>{\bf elseif} accent $=$  complex {\bf then} {\it sentential}\>(7) \\
\>{\bf elseif} position in intermediate phrase $=$ $\neg$first {\bf then} {\it sentential}\>\>(8) \\
\\
\underline{Textual Model}:\\ \\
{\bf if} preceding orthography $=$ true {\bf then}  {\it discourse} \>\>\>(9)\\
{\bf elseif} preceding orthography $=$ false {\bf then} {\it sentential} \>\>\>(10)
\end{tabbing}
\caption{\label{prosody} Decision tree representation of the manually derived classification models of Hirschberg and Litman.} 
\end{center}
}
\end{figure}

{\it Prosody} was described using Pierrehumbert's theory of English
intonation~\cite{Pierrehumbert80}.  In Pierrehumbert's theory,
intonational contours are described as sequences of low (L) and high
(H) {\it tones} in the {\it fundamental frequency (F0) contour} (the
physical correlate of pitch). Intonational contours have as their
domain the intonational phrase.  A finite-state grammar describes the
set of tonal sequences for an intonational phrase.  A well-formed {\it
intonational phrase} consists of one or more intermediate phrases
followed by a boundary tone. A well-formed {\it intermediate phrase}
has one or more pitch accents followed by a phrase accent.  {\it
Boundary tones} and {\it phrase accents} each consist of a single
tone, while {\it pitch accents} consist of either a single tone or a
pair of tones.  There are two simple pitch accents (H* and L*) and 
four {\it complex} accents (L*+H, L+H*, H*+L, and H+L*).  The * indicates
which tone is aligned with the stressed syllable of the associated
lexical item.  Note that not every stressed syllable is accented.
Lexical items that bear pitch accents are called {\it accented}, while
those that do not are called {\it deaccented}.

Prosody was manually transcribed by Hirschberg by examining the fundamental frequency
(F0) contour, and by listening to the recording.  This
transcription process was performed separately from the process of discourse/sentential
classification.
To produce the F0
contour, the recording of the corpus was digitized and pitch-tracked
using speech analysis software. This resulted in a display of the F0
where the x-axis represented time and the y-axis represented frequency
in Hz.  Various phrase final characteristics (e.g., phrase accents,
boundary tones, as well as pauses and syllable lengthening) helped to
identify intermediate and intonational phrases, while peaks or valleys
in the display of the F0 contour helped to identify pitch accents.
Similar manual transcriptions of prosodic phrasing and accent have been shown to be reliable across
coders~\cite{tobi}.

Once prosody was coded, Hirschberg and Litman represented every cue phrase in terms of the following
prosodic features.\footnote{Only the features used in Figure~\ref{prosody} are
discussed here.}
{\it Accent} corresponded to the pitch
accent (if any) that was associated with 
the cue phrase.  For both the intonational and intermediate phrases
containing each cue phrase, the feature {\it composition of phrase} represented whether
or not the cue phrase was {\it alone} in the phrase (the phrase contained only
the cue phrase, or only the cue phrase and other
cue phrases). {\it Position in phrase} represented
whether the cue phrase was {\it first} (the first lexical item in the prosodic phrase unit
-- possibly preceded by other cue phrases) or not.

The {\it textual features} used in the multiple cue phrase
study~\cite{hl93,Litman90} were extracted automatically from the
transcript.  The {\it part of speech} of each cue phrase was obtained
by running a program for tagging words with one of approximately 80
parts of speech~\cite{church88} on the transcript.\footnote{Another
syntactic feature - dominating constituent - was obtained by running
the parser Fidditch~\cite{Hindle} on the
transcript. However, since this feature did not appear in any models
manually derived from the training data~\cite{Litman90}, 
the feature was not pursued.} Several characteristics of the cue phrase's immediate
context were also noted, in particular, whether it was immediately
preceded or succeeded by {\it orthography} (punctuation or a paragraph
boundary), and whether it was immediately preceded or succeeded by a
lexical item corresponding to {\it another cue phrase}.

With this background, the classification models shown in
Figure~\ref{prosody} can now be explained.  The prosodic model
uniquely classifies any cue phrase using the features {\it composition
of intermediate phrase}, {\it position in intermediate phrase}, and
{\it accent}. When a cue phrase is uttered as a single intermediate
phrase -- possibly with other cue phrases (i.e., line (1) in
Figure~\ref{prosody}), or in a larger intermediate phrase with an
initial position (possibly preceded by other cue phrases) and a L*
accent or deaccented, it is classified as {\it discourse}.  When part
of a larger intermediate phrase and either in initial position with a
H* or complex accent, or in a non-initial position, it is {\it
sentential}.  The textual model classifies cue phrases using only the
single feature {\it preceding orthography}.\footnote{A classification
model based on part-of-speech was also developed~\cite{Litman90,hl93};
however, it did not perform as well as the model based on orthography
(the error rate of the part-of-speech model was
36.1\% in the larger test set, as opposed to 19.9\% for the
orthographic model).  
Furthermore, a model that combined orthography and
part-of-speech performed comparably to the simpler orthographic model~\cite{hl93}.
Hirschberg and Litman also had preliminary
observations suggesting that adjacency of cue phrases might prove
useful.}
When a cue phrase is preceded by any type of
orthography, it is classified as {\it discourse}; otherwise, the cue phrase is classified as {\it sentential}.

When the prosodic model was used to classify each cue phrase in
its training data, i.e., the 100 examples of ``now'' from which the model was developed,
the error rate was 2.0\%.\footnote{Following \shortciteA{hl93}, the
original 48- and 52-example sets~\cite{Hirschberg87Now} are combined.} 
The error rate of the textual model on the training examples from the
multiple cue phrase corpus was 10.6\%~\cite{Litman90}.

\begin{table*}[tb]
{\scriptsize
\begin{center}
\begin{tabular}{|l | r| r|} \hline\hline
Model	&Classifiable Cue Phrases (N=878)	&Classifiable Non-Conjuncts (N=495) \\ \hline\hline
Prosodic       	&24.6 $\pm$ 3.0 	&14.7 $\pm$ 3.2 	\\ \hline 
Textual		&19.9 $\pm$ 2.8   	&16.1 $\pm$ 3.4 	\\ \hline\hline
Default Class   &38.8 $\pm$ 3.2   	&40.8 $\pm$ 4.4 	\\ \hline
\end{tabular}
\caption{\label{hl93error} 95\% confidence intervals for the error rates (\%) of the manually derived classification models of Hirschberg and Litman, testing data (multiple cue phrase corpus).}
\end{center}
}
\end{table*}
The prosodic and textual models were evaluated by quantifying their
performance in correctly classifying example cue phrases in two test
sets of data, as shown in the rows labeled ``Prosodic'' and
``Textual'' in Table~\ref{hl93error}.  Each test set is a subset of
the 953 examples from the multiple cue phrase corpus.  The first test
set (878 examples) consists of only the {\it classifiable cue phrases},
i.e., the cue phrases that both Hirschberg and Litman classified as
{\it discourse} or that both classified as {\it sentential}.  Note
that those cue phrases that Hirschberg and Litman classified as {\it ambiguous} or that
they were unable to agree upon are not included in the classifiable
subset. (These cue phrases will be considered in the learning experiments described in
Section~\ref{exptfour}, however.)
The second test set, the {\it classifiable non-conjuncts} (495
examples), was created from the classifiable cue phrases by removing all
instances of ``and'', ``or'' and ``but''. This subset was considered
particularly reliable since 97.2\% of non-conjuncts were classifiable
compared to 92.1\% of all example cue phrases.  The error rate of the prosodic model
was 24.6\% for the classifiable cue phrases and 14.7\% for the classifiable
non-conjuncts~\cite{hl93}.  The error rate of the textual model was
19.9\% for the classifiable cue phrases and 16.1\% for the classifiable
non-conjuncts~\cite{hl93}. The last row of the table shows 
error rates for a simple ``Default Class'' baseline
model that always predicts the most frequent class in the corpus ({\it
sentential}).  These rates are 38.8\% for the classifiable cue phrases and
40.8\% for the classifiable non-conjuncts.

Although not computed by Hirschberg and Litman, Table~\ref{hl93error} also
associates margins of errors
with each error
percentage, which are used to compute confidence
intervals~\cite{stats}.  
(The margin of error is $\pm$ 2
standard errors for a 95\% 
confidence interval using a normal table.)
The lower bound of a confidence interval is
computed by subtracting the margin of error from the error rate, while
the upper bound is computed by adding the margin of error.  Thus, the
95\% confidence interval for the prosodic model on the classifiable
cue phrase test set is (21.6\%, 27.6\%).  Analysis of the confidence
intervals indicates that the improvement of both the prosodic and
textual models over the default model is significant.  
For example,
the upper bounds of the error rates of
the prosodic and textual models on the classifiable cue phrase test set -
27.6\% and 22.7\% - are both lower than the lower bound of the default
class error rate - 35.6\%.  This methodology of using statistical
inference to determine whether differences in error rates are
significant is discussed more fully in Section~\ref{evalmethod}.
\section{Experiments using Machine Learning}
\label{ml}

This section describes experiments that use the machine learning
programs C4.5~\cite{Quinlan93} and {\sc
cgrendel}~\cite{CohenIMLC92,CohenIJCAI93} to automatically induce cue
phrase classification models.  {\sc cgrendel} and C4.5 are similar to
each other and to other learning methods such as neural networks and
{\sc cart}~\cite{brieman84} in that all induce classification models from
preclassified examples.  Each program takes the following inputs:
names of the classes to be learned, names and possible values of a
fixed set of features, and the training data (i.e., a set of examples
for which the class and feature values are specified).  The output of
each program is a classification model, expressed in C4.5 as a
decision tree and in {\sc cgrendel} as an ordered set of if-then
rules.  Both {\sc cgrendel} and C4.5 learn the classification models
using greedy search guided by an ``information gain'' metric.

The first group of machine learning experiments replicate the training
and testing conditions used by  \shortciteA{hl93} (reviewed
in the previous section), to support a direct comparison of
the manual and machine learning approaches.  The second group of
experiments evaluate the utility of training from larger amounts of
data than was feasible for the manual analysis of Hirschberg and
Litman.  The third set of experiments allow the machine learning
algorithms to distinguish among the 34 cue phrases, to
evaluate the utility of developing classification models specialized
for particular cue phrases.  The fourth set of experiments consider
all the examples in the multiple cue phrase corpus, not just the
classifiable cue phrases. This set of experiments attempt to predict a
third classification {\it unknown}, as well as the classifications {\it
discourse} and {\it sentential}.  Finally, within each
of these four sets of experiments, each individual experiment learns a
classification model using a different feature representation of the
training data.  Some experiments consider features in isolation, to
comparatively evaluate the utility of each individual feature for
classification.  Other experiments consider linguistically motivated
sets of features, to gain insight into feature interactions.

\subsection{The Machine Learning Inputs}

This section describes the inputs to both of the machine learning
programs, namely, the names of the classifications to be learned, the names and
possible values of a fixed set of features, and training data
specifying the class and feature values for each example in the training set.

\subsubsection{Classifications}

The first input to each learning program specifies the names of a
fixed set of {\it classifications}.
Hirschberg and Litman's
3-way classification of cue phrases by 2
judges~\cite{hl93}  is transformed into the classifications used by the machine learning programs as shown in
Table~\ref{judge}.  
\begin{table*}[tp]
\scriptsize{
\begin{center}
\begin{tabular}{||l||r||r|r||r|r|r|r|r|r|r||} \hline\hline
& Total &
\multicolumn{2}{c||}{Classifiable Cue Phrases}& \multicolumn{7}{c||}{} \\ \hline
Classification&&\multicolumn{1}{c|}{Discourse} & \multicolumn{1}{c||}{Sentential}& \multicolumn{7}{c||}{Unknown} \\ \hline
Judge1/Judge2 	&	&D/D	&S/S	&?/?	&D/S	&S/D	&D/?	&S/?	&?/D	&?/S\\ \hline
All Cue Phrases &953	&341	&537	&59	&5 	&0	&0	&0	&5 	&6\\ \hline
Non-Conjuncts	&509    &202 	&293	&11	&1 	&0	&0	&0	&0	&2\\ \hline
\end{tabular}
\caption{\label{judge} Determining the classification of cue phrases.}
\end{center}}
\end{table*}
Recall from Section~\ref{data} that each judge classified each cue phrase
as {\it discourse}, {\it sentential}, or {\it ambiguous};
these classifications are shown as D, S, and ? in
Table~\ref{judge}.  
As
discussed in Section~\ref{data}, the {\it classifiable cue phrases} are those
cue phrases that the judges both classified as either discourse or as sentential usages.
Thus, in the machine learning experiments, a cue phrase is assigned the classification {\it
discourse} if both judges classified it as discourse (D/D, as
shown in column 3 of Table~\ref{judge}).  Similarly, a cue phrase is assigned the
classification {\it sentential} if both judges classified it as
sentential (S/S, as shown in column 4).  878 (92.1\%) of the 953
examples in the full corpus were classifiable, while 495 (97.2\%) of the
509 non-conjuncts were classifiable.

For some of the machine learning experiments, a third cue phrase classification will
also be considered.  In particular, a cue phrase is assigned
the classification {\it unknown} if both Hirschberg and Litman
classified it as {\it ambiguous} (?/?, as shown in column 5), or
if they were unable to agree upon its classification (D/S,
S/D, D/?, S/?, ?/D, ?/S, as shown in columns
6-11).  
In the full corpus, 59 cue phrases
(6.2\%) were judged ambiguous by both judges (?/?). 
There were only 5 cases
(.5\%) of true disagreement (D/S).
11 
cue phrases (1.2\%)  were judged ambiguous by the first judge 
but classified by the second judge (?/D and  ?/S).
When the conjunctions ``and,'' ``or'' and ``but'' were removed from the
corpus, only 11 examples (2.2\%) were judged ambiguous by both judges: 3
instances of ``actually,'' 2 instances each of ``because'' and
``essentially,'' and 1 instance of ``generally,'' ``indeed,''
``like'' and ``now.''  There was only 1 case (.2\%) of true disagreement (an instance of ``like''). 2 cue phrases (.4\%) - an instance each of
``like'' and ``otherwise'' - were judged ambiguous by the first judge.

\subsubsection{Features}

A second component of the input to each learning program specifies
the names and potential values of a fixed set of {\it features}.  
The set of primitive features considered in the learning
experiments are shown in Figure~\ref{features}.
\begin{figure}[tb]
{\scriptsize
\begin{center}
\begin{itemize}
\item
{\bf Prosodic Features}
\begin{itemize}
\item
length of intonational phrase (P-L): integer.
\item
position in intonational phrase (P-P): integer.
\item
length of intermediate phrase (I-L): integer.
\item
position in intermediate phrase (I-P): integer.
\item
composition of intermediate phrase (I-C): only, only cue phrases,
other.
\item
accent (A): H*, L*, L*+H, L+H*, H*+L, H+L*, deaccented, ambiguous.
\item
accent* (A*): H*, L*, complex, deaccented, ambiguous.
\end{itemize}
\item
{\bf Textual Features}
\begin{itemize}
\item
preceding cue phrase (C-P): true, false, NA.
\item
succeeding cue phrase (C-S): true, false, NA.
\item
preceding orthography (O-P): comma, dash, period, paragraph, false,
NA.
\item
preceding orthography* (O-P*): true, false, NA.
\item
succeeding orthography (O-S): comma, dash, period, false, NA.
\item
succeeding orthography* (O-S*): true, false, NA.
\item
part-of-speech (POS): article, coordinating conjunction, cardinal
numeral, subordinating conjunction, preposition, adjective, singular
or mass noun, singular proper noun, intensifier, adverb, verb base
form, NA.
\end{itemize}
\item
{\bf Lexical Feature}
\begin{itemize}
\item
token (T): actually, also, although, and, basically, because, but,
essentially, except, finally, first, further, generally, however,
indeed, like, look, next, no, now, ok, or, otherwise, right, say,
second, see, similarly, since, so, then, therefore, well, yes.
\end{itemize}
\end{itemize}
\caption{\label{features} Representation of features, for use by C4.5 and {\sc cgrendel}.}
\end{center}
}
\end{figure}
Feature values can either be a numeric value or one of a fixed set of
user-defined symbolic values.  The feature representation shown here
follows the representation of Hirschberg and Litman except as noted.
{\it Length of intonational phrase} (P-L) and {\it length of
intermediate phrase} (I-L) represent the number of words in the
intonational and intermediate phrases containing the cue phrase,
respectively. This feature was not coded in the ``now'' data, but was
coded (although not used) in the later multiple cue phrase data.  {\it Position in intonational phrase} (P-P) and {\it
position in intermediate phrase} (I-P) use numeric values rather than
the earlier symbolic values (e.g., {\it first} in
Figure~\ref{prosody}).  {\it Composition of intermediate phrase} (I-C)
replaces the value {\it alone} (meaning that the phrase contained only
the example cue phrase, or only the example plus other cue phrases)
from Figure~\ref{prosody} with the more primitive values {\it only}
and {\it only cue phrases} (whose disjunction is equivalent to {\it
alone}); I-C also uses the value {\it other} rather than 
$\neg${\it alone} (as was used in Figure~\ref{prosody}).  {\it Accent} (A) uses
the value {\it ambiguous} to represent all cases where the prosodic
analysis yields a disjunction (e.g., ``H*+L or H*'').  {\it Accent*}
(A*) re-represents some of the symbolic values of the feature {\it
accent} (A) using a more abstract level of description.  In
particular, L*+H, L+H*, H*+L, and H+L* are represented as separate
values in A but as a single value -- the superclass {\it complex} --
in A*.  While useful abstractions can often result from the learning
process, A* is explicitly represented in advance as it is a prosodic
feature representation that has the potential to be automated (see
Section~\ref{utility}). 

In all the textual features, the value {\it NA} (not applicable)
reflects the fact that 39 recorded examples were not included in the
transcription, which was done independently of the studies performed by~\shortciteA{hl93}.
In the coding used by Hirschberg and Litman, {\it preceding cue phrase} (C-P) and {\it succeeding
cue phrase} (C-S) represented the actual cue phrase (e.g., ``and'')
when there was a preceding or succeeding cue phrase; here the value
{\it true} encodes all such cases.  As with the prosodic feature set
A*,  {\it
preceding orthography*} (O-P*) and {\it succeeding orthography*}
(O-S*) re-represent some of the symbolic values of {\it preceding
orthography} (O-P) and {\it succeeding orthography} (O-S),
respectively, using a more abstract level of description (e.g., {\it
comma}, {\it dash}, and {\it period} are represented as separate
values in O-S but as the single value {\it true} in O-S*).
This is done because the reliability
of coding detailed transcriptions of orthography is not known.  
{\it Part-of-speech} (POS) represents the part of speech assigned to each
cue phrase by Church's program for tagging part of speech in
unrestricted text~\cite{church88}; while the program can assign
approximately 80 different values, only the subset of values that were
actually assigned to the cue phrases in the transcripts of the corpora
are shown in the figure.  Finally, the lexical feature {\it token} (T)
is new to this study, and represents the actual cue phrase being
described.

\subsubsection{Training Data}

The final input to each learning program is {\it training data},
i.e., a set of examples for which the class and feature values are
specified.  Consider the following utterance, taken from the multiple cue phrase corpus~\cite{hl93}:
\begin{description}
\item[Example 1] [({\it Now}) ({\it now} that we have all been welcomed here)] it's time to get on with
the business of the conference.
\end{description}
This utterance contains two cue phrases, corresponding to the two
instances of ``now''.  The brackets and parentheses illustrate the
intonational and intermediate phrases, respectively, that contain the
example cue phrases.  Note that a single intonational phrase contains both examples,
but that each example is uttered in a different intermediate phrase.  If
we were only interested in the feature {\it length of intonational
phrase} (P-L), the two examples would be represented in the training
data as follows:
\begin{flushleft}
\begin{tabular}{r l}
P-L	&Class	\\ \hline 9	&discourse\\ 9	&sentential\\ 
\end{tabular}
\end{flushleft}
The first column indicates the value assigned to the feature P-L,
while the second column indicates how the example was classified.
Thus, the length of the intonational phrase containing the first
instance of ``now'' is 9 words, and the example cue phrase is classified as a
discourse usage.  If we were only interested in the feature {\it composition of
intermediate phrase} (I-C), the two examples would instead be represented
in the training data as follows:
\begin{flushleft}
\begin{tabular}{l l}
I-C	&Class	\\ \hline 
only	&discourse\\ 
other	&sentential\\ 
\end{tabular}
\end{flushleft}
That is, the intermediate phrase containing the first instance of ``now''
contains only the cue phrase ``now'', while the intermediate phrase
containing the second instance of ``now'' contains ``now'' as well as
7 other lexical items that are not cue phrases.  Note that while the value of P-L
is the same for both examples, the value of I-C is different. 

\subsection{The Machine Learning Outputs}

The output of both machine learning programs are {\it classification
models}.  In C4.5 the model is expressed as a {\it decision tree},
which consists of either a leaf node (a class assignment), or a
decision node (a test on a feature, with one branch and subtree for
each possible outcome of the test).  The following example illustrates
the non-graphical representation for a decision node testing a feature with n
possible values: 
\begin{tabbing}
{\bf if} \=(\= sss \= sss \= ssssss\kill
{\bf if} test$_{1}$ {\bf then} \ldots \\
\ldots\\
{\bf elseif} test$_{n}$ {\bf then} \ldots
\end{tabbing}
Tests
are of the form ``feature operator value''\footnote{An additional type
of test may be invoked by a C4.5 option.}. ``Feature'' is the name of a
feature (e.g. {\it accent}), while
``value'' is a valid value for that feature (e.g., {\it deaccented}).  For
features with symbolic values (e.g., {\it accent}), there is one
branch for each symbolic value, and the operator ``$=$'' is used.  For
features with numeric values (e.g., {\it length of intonational
phrase}), there are two branches, each comparing the numeric value
with a threshold value; the operators ``$\leq$'' and
``$>$'' are used.  Given a decision tree, a cue phrase is
classified by starting at the root of the tree and following
the appropriate branches until a leaf is reached.
Section~\ref{results} shows example decision trees produced by C4.5.

In {\sc cgrendel} the classification model is expressed as an ordered
set of {\it if-then rules} of the following form:
\begin{tabbing}
{\bf if} test$_{1}$ $\wedge$ \ldots $\wedge$ test$_{k}$ {\bf then} {\it class}
\end{tabbing}
The ``if'' part of a rule is a conjunction of tests on the values of
(varying) features, where tests are again of the form ``feature
operator value.''  As in C4.5, ``feature'' is the name of a feature, and
``value'' is a valid value for that feature.  Unlike C4.5, the operators
$=$ or $\neq$ are used for features with symbolic values, while $\leq$
or $\geq$ are used for features with numeric values.  The ``then''
part of a rule specifies a class assignment (e.g, {\it discourse}).
Given a set of if-then rules, a cue phrase is classified using the rule
whose ``if'' part is satisfied. If there or two or more such rules
and the rules disagree on the class of an example,
{\sc cgrendel} applies one of two conflict resolution strategies
(chosen by the user): choose the first rule, or choose the rule that
is most accurate on the data.  The experiments reported here use the
second strategy.  If there are no such rules, {\sc cgrendel} assigns a
default class.  Section~\ref{results} shows example rules produced by
{\sc cgrendel}.

Both C4.5 and {\sc cgrendel} learn  their classification models using
greedy search guided by an ``information gain'' metric.  
C4.5 uses a divide and conquer process:
training examples are recursively divided into subsets (using the tests
discussed above), until all of
the subsets belong to a single class.  
The test chosen to divide the examples is
that which maximizes a metric called a gain ratio (a
local measure of progress, which does not consider any subsequent
tests); this metric is based on information theory and is discussed in
detail by \shortciteA{Quinlan93}.  Once a
test is selected, there is no backtracking.  Ideally, the set of chosen tests should
result in a small final decision tree.  {\sc cgrendel} generates its
set of if-then rules using a method called {\it separate and
conquer} (to highlight the similarity with divide and conquer):
\begin{quote}
Many rule learning systems generate hypotheses using a greedy strategy
in which rules are added to the rule set one by one in an effort to
form a small cover of the positive examples; each rule, in turn is
created by adding one condition after another to the antecedent until
the rule is consistent with the negative data.~\cite{CohenIJCAI93}
\end{quote}

Although {\sc cgrendel} is claimed to have two advantages over C4.5,
these advantages do not come into play for
the experiments reported here. First, if-then rules appear to be
easier for people to understand than decision trees~\cite{Quinlan93}.
However, for the cue phrase classification task, the decision trees
produced by C4.5 are quite compact and thus easily understood.
Furthermore, a rule representation can be derived from C4.5 decision
trees, using the program C4.5rules.  Second, {\sc cgrendel} allows
users to exploit prior knowledge of a learning problem, by
constraining the syntax of the rules that can be learned.  However, no
prior knowledge is exploited in the cue phrase experiments.  The main
reason for using both C4.5 and {\sc cgrendel} is to  increase
the reliability of any comparisons between the machine learning and manual
results.  In particular, if comparable results are obtained using both
C4.5 and {\sc cgrendel}, then any performance differences between the
learned and manually derived classification models are less likely to
be due to the specifics of a particular learning program,
and more likely to reflect the learned/manual distinction.

\subsection{Evaluation}
\label{evalmethod}

The output of each machine learning experiment is a classification
model that has been learned from the training data.  These learned
models are qualitatively evaluated by examining their linguistic
content, and by comparing them with the manually derived models of
Figure~\ref{prosody}.  The learned models are also quantitatively
evaluated by examining their error rates on testing data and by comparing these error
rates to each other and to the error rates shown in Table~\ref{hl93error}.  The {\it
error rate} of a classification model is computed by using the model
to predict the classifications for a set of examples where the
classifications are already known, then comparing the predicted and
known classifications.  In the cue phrase domain, the error rate is
computed by summing the number of discourse examples misclassified as
sentential with the number of sentential examples misclassified as
discourse, then dividing by the total number of examples.

The error rates of the learned classification models are estimated using two
methodologies.  {\it Train-and-test error rate
estimation}~\cite{KulikowskiBook90} ``holds out'' a test set of
examples, which are not seen until after training is completed.  That
is, the model is developed by examining only the training examples;
the error of the model is then estimated by using the model to
classify the test examples.  This was the evaluation method used
by Hirschberg and Litman.  The resampling method of {\it
cross-validation}~\cite{KulikowskiBook90} estimates error rate using
multiple train-and-test experiments. For example, in 10-fold
cross-validation, instead of dividing examples into training and test
sets once, 10
runs of the learning program are performed. The total set of examples is randomly divided
into 10 disjoint test sets; each run thus uses the 90\% of the
examples not in the test set for training and the remaining 10\% for
testing.  Note that for each iteration of the cross-validation,
the learning process begins from scratch; thus a new 
classification model is learned from each training sample.  An
estimated error rate is obtained by averaging the error rate on the
testing portion of the data from each of the 10 runs.  While this
method does not make sense for humans, computers can truly ignore
previous iterations.  For sample sizes in the hundreds (the classifiable
subset of the multiple cue phrase sample and the classifiable non-conjunct subset provide 878 and 495 examples,
respectively) 10-fold cross-validation often provides a better
performance estimate than the hold-out method~\cite{KulikowskiBook90}.
The major advantage is that in cross-validation all examples are
eventually used for testing, and almost all examples are used
in any given training run.

The best performing learned models are identified by comparing their
error rates to the error rates of the other learned models and to
the manually derived error rates.  To determine whether the fact that
an error rate E1 is lower than another error rate E2 is also significant,
statistical inference is used.  In particular, confidence intervals
for the two error rates are computed, at a 95\% confidence level. When
an error rate is estimated using only a single error rate on a test
set (i.e., the train-and-test methodology), the confidence interval is
computed using a normal approximation to the binomial
distribution~\cite{stats}.  When the error rate is estimated using the
average from multiple error rates (i.e., the cross-validation
methodology), the confidence interval is computed using a
{\it t}-Table~\cite{stats}.  If the upper bound of the 95\% confidence
interval for E1 is lower than the lower bound of the 95\% confidence
interval for the error rate E2, then the difference between E1 and E2
is assumed to be significant.\footnote{Thanks to William Cohen for
suggesting this methodology.}

\subsection{The Experimental Conditions}

This section describes the conditions used in each set of machine
learning experiments.  The experiments differ in their use of training
and testing corpora, methods for estimating error rates, and in the
features and classifications used.  The actual results of the
experiments are presented in Section~\ref{results}.

\subsubsection{Four Sets of Experiments} 

The learning experiments can be conceptually divided into four sets.
Each experiment in the first set estimates error rate using the
train-and-test method, where the training and testing samples are
those used by  \shortciteA{hl93} (the ``now'' data and the two subsets of the
multiple cue phrase corpus, respectively).  This allows
a direct comparison of the manual and machine learning approaches.
However, only the prosodic experiments conducted by \shortciteA{hl93} are replicated.
The textual training and testing
conditions are not replicated as the original training
corpus
(the first 17 minutes of the multiple cue phrase corpus)~\cite{Litman90} is a subset of, rather than disjoint from, the test corpus (the full 75 minutes of the multiple cue phrase corpus)~\cite{hl93}. 

In contrast, each experiment in the second set uses cross-validation
to estimate error rate.  
Furthermore, both training and testing samples are taken
from the multiple cue phrase corpus.
Each experiment uses 90\% of
the examples from the multiple cue phrase data for training, and the
remaining 10\% for testing.  Thus each experiment in the second set
trains from much larger amounts of data (790 classifiable examples, or 445 classifiable non-conjuncts)
than each experiment in the first set (100 ``nows'').
The reliability of the testing is not compromised due to the use of cross-validation~\cite{KulikowskiBook90}.

Each experiment in the third set 
replicates an experiment in the second set, with the exception that
the learning program is now allowed to distinguish between cue
phrases.  This is done by adding a feature
representing the cue phrase (the feature {\it token} from Figure~\ref{features}) to each experiment from the second set.  Since the potential use of such a lexical feature
was noted but not used by \shortciteA{hl93}, these experiments provide
qualitatively new linguistic insights into the data.  For example, 
the same features may now be used differently to predict the classifications of different cue phrases
or sets of cue phrases.

Finally, each experiment in the fourth set replicates an experiment in
the first, second, and third set, with the exception that all 953
examples in the multiple cue phrase corpus are now considered.
This is because in practice,
any learned cue phrase classification model will likely be used to
classify all cue phrases, even those that are difficult for human
judges to classify.
The experiments in the fourth set allow
the learning programs to attempt to learn the class {\it unknown}, in
addition to the classes {\it discourse} and {\it sentential}.

\subsubsection{Feature Representations within Experiment Sets} 

Within each of these four sets of experiments, each
individual experiment represents the data using a different subset of
the available features.  
First, the data is represented in each of 14 {\it single
feature sets}, corresponding to each prosodic and textual
feature shown in Figure~\ref{features}.  
These experiments comparatively evaluate the utility of each individual
feature for classification.  
The representations of Example 1 shown above illustrate how data is represented
using the single feature set P-L, and using the single feature set I-C.

Second, the data is represented in each of the 13 {\it multiple feature sets} shown
in Table~\ref{samples}.
\begin{table*}[tb]
\scriptsize{
\begin{center}
\begin{tabular}{|l| r r r r r r r| r r r r r r r|} \hline\hline
		&P-L	&P-P	&I-L	&I-P	&I-C	&A &A*	&C-P	&C-S	&O-P &O-P* &O-S &O-S*	&POS	\\ \hline\hline
prosody	     	&X	&X	&X	&X	&X	&X&X	&	&	&&	&&	&	\\ \hline
hl93features  	&  	&	&	&X	&X	&X&X	&	&	&&	&&	&	\\ \hline
phrasing     	&X	&X	&X	&X	&X	&&	&	&	&&	&&	&	\\ \hline
length	     	&X	&	&X	&	&	&&	&	&	&&	&&	&	\\ \hline
position     	&	&X	&	&X	&	&&	&	&	&&	&&	&	\\ \hline
intonational 	&X	&X	&	&	&	&&	&	&	&&	&&	&	\\ \hline
intermediate	&	&	&X	&X	&X	&&	&	&	&&	&&	&	\\ \hline\hline
text	     	&	&	&	&	&	&&	&X	&X	&X&X	&X&X	&X	\\ \hline
adjacency    	&	&	&	&	&	&&	&X	&X	&&	&&	&	\\ \hline
orthography  	&	&	&	&	&	&&	&	&	&X&X	&X&X	&	\\ \hline
preceding    	&	&	&	&	&	&&	&X	&	&X&X	&&	&	\\ \hline
succeeding   	&	&	&	&	&	&&	&	&X	&&	&X&X	&	\\ \hline\hline
speech-text  	&X	&X	&X	&X	&X	&X&X	&X	&X	&X&X	&X&X	&X	\\ \hline
\end{tabular}
\caption{\label{samples} Multiple feature sets and their components.}
\end{center}}
\end{table*}
Each of these sets contains a linguistically motivated subset of at
least 2 of the 14 features.  The first 7 sets use
only prosodic features.  {\it Prosody} considers all the prosodic
features that were coded for each example cue phrase. {\it Hl93features} considers
only the coded features that were also used in the model shown in
Figure~\ref{prosody}.  {\it Phrasing} considers all features of both the
intonational and intermediate phrases containing the example cue phrase (i.e.,
length of phrase, position of example in phrase, and composition of
phrase).  {\it Length} and {\it position} each consider only one of these
features, but with respect to both the intonational and intermediate
phrase.  Conversely, {\it intonational} and {\it intermediate} each
consider only one type of phrase, but consider all of the features.
The next 5 sets use only textual features.  {\it Text} considers all the
textual features.  {\it Adjacency} and {\it orthography} each consider a
single textual feature, but consider both the preceding and succeeding
immediate context.  {\it Preceding} and {\it succeeding}
consider contextual features relating to both orthography and cue
phrases, but limit the context.  The last set, {\it speech-text}, uses all of the prosodic and textual features.

Figure~\ref{examples} illustrates how the two example cue phrases in Example 1
would be represented using {\it speech-text}.
\begin{figure}[tb]
{\scriptsize
\begin{center}
\begin{description}
\item[Example 1] [({\it Now}) ({\it now} that we have all been welcomed here)] it's time to get on with
the business of the conference.
\end{description}
\medskip
\begin{tabular}{ r r r r l l l l l l l l l l l} 
P-L	&P-P	&I-L	&I-P	&I-C	&A	&A*	&C-P	&C-S	&O-P	&O-P*	&O-S	&O-S*	&POS	&Class\\ \hline
9	&1	&1	&1	&only	&H*+L	&complex&f	&t	&par. 	&t	&f	&f	&adv.	&disc.\\ 
9	&2	&8	&1	&other	&H*	&H*	&t	&f	&f	&f	&f	&f	&adv.	&sent. \\ 
\end{tabular}
\caption{\label{examples} Representation of Example 1 in feature set speech-text.}
\end{center}}
\end{figure}
Consider the feature values for the first example cue phrase.  Since this example is
the first lexical item in both the intonational and intermediate
phrases which contain it, its position in both phrases (P-P and
I-P) is 1.  Since the intermediate phrase containing the cue phrase
contains no other lexical items, its length (I-L) is 1 word and its
composition (I-C) is {\it only} the cue phrase.  The values for A and
A* indicate that when the intonational phrase is described as a
sequence of tones, the complex pitch accent H*+L is associated
with the cue phrase.  With respect to the textual features, the utterance was transcribed such that it
began a new paragraph. Thus the example cue phrase was not preceded by another cue
phrase (C-P), but it was preceded by a form of orthography (O-P and
O-P*).  Since the example cue phrase was immediately followed by
another instance of ``now'' in the transcription, the cue phrase was
succeeded by another cue phrase (C-S) but was not succeeded by
orthography (O-S and O-S*).
Finally, the output of the part of speech
tagging program when run on the transcript of the corpus yields the value {\it adverb} for the cue phrase's part of speech (POS).

The first set of experiments replicate only the prosodic experiments conducted by  \shortciteA{hl93}; 
cue phrases are represented using the subset of the
feature sets that only consist of prosodic features. 
In the second set of experiments, examples are represented
using all 27 different feature sets (the 14 single feature sets and the 13 multiple feature sets).  
In the third set of experiments, examples are represented using 27 {\it tokenized feature
sets}, constructed by adding the lexical feature {\it token} from Figure~\ref{features} (the cue phrase being
described) to each of the 14 single and 13 multiple feature sets from
the second set of experiments.
These tokenized feature sets will be referred to using the names of the single and
multiple feature sets, concatenated with ``+''.
The following illustrates how the two cue phrases in Example 1 would be represented 
using P-L+:
\begin{flushleft}
\begin{tabular}{r l l}
P-L	&T &Class	\\ \hline 
9	&now &discourse\\ 
9	&now &sentential\\
\end{tabular}
\end{flushleft}
The representation is similar to the P-L representation shown
earlier, except for the second column which indicates the value
assigned to the feature {\it token} (T).

\section{Results}
\label{results}

This section examines the results of running the two learning programs
-- C4.5 and {\sc cgrendel} -- in the four sets of cue phrase
classification experiments described above.  The
learned classification models will be compared with the classification
models shown in Figure~\ref{prosody}, while the error rates of the
learned classification models will be compared with the error rates
shown in Table~\ref{hl93error} and with the error rates of the other learned models.  As will be seen, the results suggest
that machine learning is useful for automating the generation of
linguistically viable classification classification models, for
generating classification models that perform with lower error rates
than manually developed hypotheses, and for adding to the body of
linguistic knowledge regarding cue phrases.

\subsection{Experiment Set 1: Replicating Hirschberg and Litman}

The first group of experiments replicate the training, testing, and
evaluation conditions used by \shortciteA{hl93}, in order to investigate how well
machine learning performs in comparison to the manual development of
cue phrase classification models.  

\begin{figure}
{\scriptsize
\begin{center}
\begin{tabbing}
ssss \= ssss \= sssssssssssssssssssssssssssssssssssssssssssssssssssssssssssssssssssssssss\= \kill
\underline{Manually derived prosodic model (repeated from Figure~\ref{prosody})}:\\ \\
{\bf if} composition of intermediate phrase $=$ alone {\bf then}  {\it discourse} \>\>\>(1) \\
{\bf elseif}  composition of intermediate phrase $=$ $\neg$alone {\bf then} \>\>\>(2) \\  
\>{\bf if} position in intermediate phrase $=$ first {\bf then}\>\>(3)\\
\>	\>{\bf if} accent $=$  deaccented {\bf then}  {\it discourse} \>(4) \\ 
\>	\>{\bf elseif} accent $=$   L* {\bf then}  {\it discourse}\>(5) \\
\>	\>{\bf elseif} accent $=$   H* {\bf then}  {\it sentential}\>(6) \\
\>	\>{\bf elseif} accent $=$  complex {\bf then} {\it sentential}\>(7) \\
\>{\bf elseif} position in intermediate phrase $=$ $\neg$first {\bf then} {\it sentential}\>\>(8) \\
\\
\\
\underline{Decision tree learned from prosody, from phrasing, and from position using C4.5:}\\ \\
{\bf if} position in intonational phrase $=$ first {\bf then} {\it
discourse} \\ {\bf elseif} position in intonational phrase $=$ $\neg$first
{\bf then} \\ 	\>{\bf if} position in intermediate phrase $=$ first
{\bf then} {\it discourse} \\ 	\>{\bf elseif} position in
intermediate phrase $=$ $\neg$first {\bf then} {\it sentential} \\
\\
\underline{Ruleset learned from prosody, from phrasing, and from position using CGRENDEL}:\\ \\
{\bf if} (position in intonational phrase $\neq$ first) $\wedge$
(position in intermediate phrase $\neq$ first) {\bf then} {\it
sentential} \\ default is on {\it discourse}\\
\\
\\
\underline{Decision tree learned from P-P and from intonational using C4.5:}\\ \\
{\bf if} position in intonational phrase $=$ first {\bf then} {\it
discourse} \\ {\bf elseif} position in intonational phrase $=$
$\neg$first {\bf then} {\it sentential} \\
\\
\underline{Ruleset learned from P-P and from intonational using CGRENDEL}:\\ \\
{\bf if} position in intonational phrase $\neq$ first {\bf then} {\it
sentential} \\ default is on {\it discourse}\\
\end{tabbing}
\caption{\label{output-now} Example C4.5 and {\sc cgrendel} classification models learned from different prosodic feature representations of the ``now'' data.}
\end{center}}
\end{figure}
Figure~\ref{output-now} shows the best performing prosodic classification
models learned by the two machine learning programs; the top of the
figure replicates the manually derived prosodic model from
Figure~\ref{prosody} for ease of comparison.  
When all of
the prosodic features are used to represent the 100 training examples of
``now'' (i.e., each example is represented using feature set {\it
prosody} from Table~\ref{samples})\footnote{In Experiment Set 1, the
feature set {\it prosody} does not contain the features P-L and I-L.
Recall that phrasal length was only coded in the later multiple cue
phrase study.}, the classification models that are learned are shown
after the manually derived model at the top of
Figure~\ref{output-now}.  Note that using both learning programs, the
same decision tree is also learned when the smaller feature sets
{\it phrasing} and {\it position} are used to represent the ``now'' data.
The bottom portion of the figure shows the classification models that
are learned when the same examples are represented using only the
single prosodic feature {\it position in intonational phrase} (P-P);
the same model is also learned when the examples are represented using
the multiple feature set {\it intonational}.

Recall that C4.5 represents each learned classification model as a
decision tree.  Each level of the tree (shown by indentation) specifies a test on a single
feature, with a branch for every possible outcome of the test. A
branch can either lead to the assignment of a class, or to another
test.  For example, the C4.5 classification model learned from {\it
prosody} classifies cue phrases using the two features {\it position
in intonational phrase} (P-P) and {\it position in intermediate
phrase} (I-P).  Note that not all of the available features in {\it
prosody} (recall Table~\ref{samples}) are used in the decision tree.
The tree initially branches based on the value of the feature {\it
position in intonational phrase}.\footnote{For ease of comparison to
Figure~\ref{prosody}, the original symbolic representation of the
feature value is used rather than the integer representation shown in
Figure~\ref{features}.} The first branch leads to the class assignment
{\it discourse}.  The second branch leads to a test of the feature
{\it position in intermediate phrase}.  The first branch of this test
leads to the class assignment {\it discourse}, while the second branch
leads to {\it sentential}.  C4.5 produces both unsimplified and pruned
decision trees.  The goal of the pruning process is to
take a complex decision tree that may also be overfitted to the
training data, and to produce a tree that is more comprehensible and
whose accuracy is not comprised~\cite{Quinlan93}.  
Since almost all trees are improved by pruning~\cite{Quinlan93},
only simplified decision trees are considered in this paper.

In contrast, {\sc cgrendel} represents each learned classification
model as a set of if-then rules.  Each rule specifies a conjunction of
tests on various features, and results in the assignment of a class.
For example, the {\sc cgrendel} ruleset learned
from {\it prosody} classifies cue phrases using the two features {\it
position in intonational phrase} (P-P) and {\it position in
intermediate phrase} (I-P) (the same two features used in the C4.5
decision tree).  If the values of both features are not {\it first}, the
if-then rule applies and the cue phrase is classified as {\it
sentential}.  If the value of either feature is {\it first}, the
default applies and the cue phrase is classified as {\it discourse}.

An examination of the learned classification models of
Figure~\ref{output-now} shows that they are comparable in content to the
portion of the manually derived model that classifies cue phrases
solely on phrasal position (line (8)).  In particular, all of the
classification models say that if the cue phrase is not in an initial
phrasal position classify it as {\it sentential}.
On the other hand, the manually derived model also assigns the class
{\it sentential} given an initial phrasal position in conjunction with
certain combinations of phrasal composition and accent; the learned
classification models instead classify the cue phrase as {\it discourse} in
all other cases. As will be shown, the further discrimination of the
manually obtained model does not significantly improve performance
when compared to the learned classification models, and in fact in one
case significantly degrades performance.

The error rates of the learned classification models on the ``now''
training data from which they were developed is as follows: 6\% for
the models learned from {\it prosody}, {\it phrasing} and {\it position}, and 9\% for the models
learned from P-P and {\it intonational}.  Recall from Section~\ref{data} that the error rate
of the manually developed prosodic model of Figure~\ref{prosody} on
the same training data was 2\%.
\begin{table*}[tb]
{\scriptsize
\begin{center}
\begin{tabular}{|l | r| r|} \hline\hline
Model	&Classifiable Cue Phrases (N=878)	&Classifiable Non-Conjuncts (N=495) \\ \hline\hline 
P-P		&{\it 18.3 $\pm$ 2.6}	&16.6 $\pm$ 3.4	\\ \hline
prosody		&27.3 $\pm$ 3.0		&17.8 $\pm$ 3.4	\\ \hline
phrasing	&27.3 $\pm$ 3.0	 	&17.8 $\pm$ 3.4	\\ \hline
position	&27.3 $\pm$ 3.0		&17.8 $\pm$ 3.4	\\ \hline
intonational	&{\it 18.3 $\pm$ 2.6}	&16.6 $\pm$ 3.4	\\ \hline\hline
manual prosodic	&24.6 $\pm$ 3.0		&14.7 $\pm$ 3.2	\\ \hline
\end{tabular}
\caption{\label{nowtesting} 95\%-confidence intervals for the error rates (\%) of the best performing {\sc cgrendel} prosodic classification models, testing data. (Training data was the ``now'' corpus; testing data was the multiple cue phrase corpus.)}
\end{center}
}
\end{table*}

Table~\ref{nowtesting} presents 95\% confidence intervals for the error
rates of the best performing {\sc cgrendel} prosodic classification
models.  For ease of
comparison, the row labeled ``manual prosodic'' presents the error
rates of the manually developed prosodic model of Figure~\ref{prosody}
on the same two test sets, which were originally shown in
Table~\ref{hl93error}.  
The table includes all the {\sc cgrendel}
models whose performance matches or exceeds the manual performance. 

Comparison of the error rates of the learned and manually developed
models suggests that machine learning is an effective technique for
automating the development of cue phrase classification models.  In
particular, within each test set, the 95\% confidence interval for the
error rate of the classification models learned from the multiple
feature sets {\it prosody}, {\it phrasing}, and {\it position} each
overlaps with the confidence interval for the error rate of the manual
prosodic model.  This is also true for the error rates of P-P and {\it
intonational} in the classifiable non-conjunct test set.  
Thus,
machine learning supports the automatic construction of a variety of
cue phrase classification models that achieve similar performance as
the manually constructed models.

The results from P-P and from {\it intonational} in the classifiable cue phrase test
set are shown in italics, as
they suggest that machine learning may also be useful for improving
performance.  Although the very simple classification model
learned from P-P and {\it intonational} performs worse than the manually derived model on the
training data, when tested on the classifiable cue phrases, the learned model
(with an upper bound error rate of 20.9\%) outperforms the manually
developed model (with a lower bound error rate of
21.6\%).  This suggests that the manually derived model might have
been overfitted to the training data, i.e.,
that the prosodic feature set most useful for classifying ``now'' did
not generalize to other cue phrases.
As noted above, the use of
simplified learned classification models helps to guard against
overfitting in the learning approach.
The ease of inducing classification models from many different sets of
features using machine learning supports the generation and evaluation
of a wide variety of hypotheses (e.g. P-P, which was a high performing
but not the optimal performing model on the training
data).

Note that the manual prosodic manual performs significantly
better in the smaller test set (which does not contain the cue phrases
``and'', ``or'', and ``but'').  In contrast, the performance
improvement for P-P and {\it intonational} in the smaller test set is
not significant.  This also suggests that the manually derived model
does not generalize as well as the learned models.

Finally, for the feature sets shown in Table~\ref{nowtesting}, the
decision trees produced by C4.5 perform with the same error
rates as the rulesets produced by {\sc cgrendel}, for both test sets.
Recall from Figure~\ref{output-now} that the C4.5 decision trees
and {\sc cgrendel} rules are in fact semantically equivalent for each 
feature set.
The fact that comparable results are obtained using C4.5 and {\sc
cgrendel} adds an extra degree of reliability to the experiments.  In
particular, the duplication of the results suggests that the ability to
match and perhaps even to improve upon manual performance by using
machine learning is not due to the specifics of either learning
program.

\subsection{Experiment Set 2: Using Different Training Sets}

The second group of experiments evaluate the utility of training from
larger amounts of data.  This is done by using 10-fold
cross-validation to estimate error, where for each run 90\% of the
examples in a sample are used for training (and over the 10 runs, all
of the examples are used for testing).  In addition, the experiments
in this second set take both the training and testing data from the
multiple-cue phrase corpus, in contrast to the previous set of
experiments where the training data was taken from the ``now'' corpus.
As will be seen, these changes improve the results, such that
more of the learned classification models perform with lower or comparable 
error rates when compared to the manually developed models.  

\subsubsection{Prosodic Models}

\begin{table*}[tb]
{\scriptsize
\begin{center}
\begin{tabular}{|l | r| r|} \hline\hline
Model	&Classifiable Cue Phrases (N=878)	&Classifiable Non-Conjuncts (N=495) \\ \hline\hline 
P-L			&33.0 $\pm$ 5.9		&(33.2 $\pm$ 1.9)		\\ \hline	
P-P			&{\it 16.1 $\pm$ 3.5}	&18.8 $\pm$ 4.2 	 	\\ \hline	
I-L			&25.5 $\pm$ 3.7		&(25.6 $\pm$ 2.8) 	 	\\ \hline	
I-P			&25.9 $\pm$ 4.9		&19.4 $\pm$ 3.1 	        \\ \hline	
I-C			&(36.5 $\pm$ 5.4)	&(35.2 $\pm$ 3.4) 	 	\\ \hline	
A			&28.6 $\pm$ 3.6		&(30.2 $\pm$ 3.1)        	\\ \hline	
A*			&28.3 $\pm$ 4.3		&(28.4 $\pm$ 1.7) 		\\ \hline \hline	
prosody			&{\it 15.5 $\pm$ 2.6}	&17.2 $\pm$ 3.1		 	\\ \hline	
hl93features		&29.4 $\pm$ 3.3	 	&18.2 $\pm$ 4.2 	 	\\ \hline 	
phrasing 		&{\it 16.1 $\pm$ 3.4}	&19.6 $\pm$ 3.9 		\\ \hline 
length			&26.1 $\pm$ 3.8		&(27.4 $\pm$ 3.4)	 	\\ \hline	 
position		&{\it 18.2 $\pm$ 2.3}	&19.4 $\pm$ 2.8 	 	\\ \hline	 
intonational		&{\it 17.0 $\pm$ 4.0}	&20.6 $\pm$ 3.6 	 	\\ \hline	 
intermediate		&21.9 $\pm$ 2.3		&19.4 $\pm$ 5.7 	 	\\ \hline \hline
manual prosodic		&24.6 $\pm$ 3.0		&14.7 $\pm$ 3.2			\\ \hline
\end{tabular}
\caption{\label{prosodic} 95\%-confidence intervals for the error rates (\%) of all
{\sc cgrendel} prosodic classification models, testing data. (Training and testing were done from the multiple cue phrase corpus using cross-validation.) }
\end{center}
}
\end{table*}
Table~\ref{prosodic} presents the error rates of the classification
models learned by {\sc cgrendel}, in the 28
different prosodic experiments. (For Experiment Sets 2 and 3, the C4.5 
error rates are presented in Appendix A.) Each numeric cell shows the 95\%
confidence interval for the error rate, which is equal to 
the error percentage obtained by cross-validation $\pm$ the margin
of error ($\pm$ 2.26 standard errors, using a {\it t}-Table). 
The top portion of the table considers the models learned from the single prosodic
feature sets (Figure~\ref{features}), the middle portion
considers the models learned from the multiple feature sets (Table~\ref{samples}),
while the last row considers the manually developed prosodic model.
The error rates shown in italics indicate that the performance of the
learned classification model exceeds the performance of the manual
model (given the same test set).  
The error rates shown in parentheses indicate the opposite
case - that the performance of the manual model exceeds the
performance of the learned model. Such cases were omitted
in Table~\ref{nowtesting}.

As in Experiment Set 1, comparison of the error rates of the learned
and manually developed models suggests that machine learning is an
effective technique for not only automating the development of cue
phrase classification models, but also for improving performance.  
When evaluated on the classifiable cue phrase test set, five
learned models have improved performance compared to the manual model;
all of the models except I-C perform
at least comparably to the manual model.  
Note that in
Experiment Set 1, only two learned models outperformed the
manual model, and only five learned models performed at least
comparably.
The ability to use large training sets 
thus appears
to be an advantage of the automated approach.

When tested on the classifiable non-conjuncts (where the error rate
of the manually derived model decreases), machine learning is useful for
automating but not for improving performance.  This might reflect the
fact that the manually derived theories already achieve optimal
performance with respect to the examined features in this less noisy
subcorpus, and/or that the automatically derived theory for this
subcorpus was based on a smaller training set than used in the larger
subcorpus.

\begin{figure}
{\scriptsize
\begin{center}
\begin{tabbing}
ssss \= ssss \= sssssssssssssssssssssssssssssssssssssssssssssssssssssssssssssssssssssssss\= \kill
\underline{Manually derived prosodic model (repeated from Figure~\ref{prosody})}:\\ \\
{\bf if} composition of intermediate phrase $=$ alone {\bf then}  {\it discourse} \>\>\>(1) \\ 
{\bf elseif}  composition of intermediate phrase $=$ $\neg$alone {\bf then} \>\>\>(2) \\  
\>{\bf if} position in intermediate phrase $=$ first {\bf then}\>\>(3)\\ 
\>	\>{\bf if} accent $=$  deaccented {\bf then}  {\it discourse} \>(4) \\ 
\>	\>{\bf elseif} accent $=$   L* {\bf then}  {\it discourse}\>(5) \\
\>	\>{\bf elseif} accent $=$   H* {\bf then}  {\it sentential}\>(6) \\
\>	\>{\bf elseif} accent $=$  complex {\bf then} {\it sentential}\>(7) \\
\>{\bf elseif} position in intermediate phrase $=$ $\neg$first {\bf then} {\it sentential}\>\>(8) \\
\\ 
\underline{Decision tree learned from P-P using C4.5}:\\ 
\\
{\bf if} position in intonational phrase $\leq$ 1 {\bf then} {\it discourse}\\ 
{\bf elseif} position in intonational phrase $>$ 1 {\bf then} {\it sentential}\\
\\
\underline{Ruleset learned from P-P using CGRENDEL}: \\ 
\\
{\bf if} position in intonational phrase $\geq$ 2 {\bf then} {\it sentential}\\ 
default is on {\it discourse}\\
\\
\underline{Decision tree learned from prosody using C4.5}: \\ 
\\
{\bf if} position in intonational phrase $\leq$ 1 {\bf then} \\ 
\>{\bf if} position in intermediate phrase $\leq$ 1 {\bf then} {\it discourse}\\
\>{\bf elseif} position in intermediate phrase $>$ 1 {\bf then} {\it sentential}\\
{\bf elseif} position in intonational phrase $>$ 1 {\bf then} 	\\
\>{\bf if} length of intermediate phrase $\leq$ 1 {\bf then} {\it discourse}\\
\>{\bf elseif} length of intermediate phrase $>$ 1 {\bf then} {\it sentential}\\
\\
\underline{Ruleset learned from prosody using CGRENDEL}:\\ 
\\
{\bf if} (position in intonational phrase $\geq$ 2) $\wedge$ (length of intermediate phrase $\geq$ 2) {\bf then} {\it sentential} \\ 
{\bf if} (7 $\geq$ position in intonational phrase $\geq$ 4) $\wedge$ (length of intonational phrase $\geq$ 10) {\bf then} {\it sentential} \\ 
{\bf if} (length of intermediate phrase $\geq$ 2) $\wedge$ (length of intonational phrase $\leq$ 7) $\wedge$ (accent $=$ H*) {\bf then} {\it sentential} \\ 
{\bf if} (length of intermediate phrase $\geq$ 2) $\wedge$ (length of intonational phrase $\leq$ 9) $\wedge$ (accent $=$ H*+L) {\bf then} {\it sentential}\\ 
{\bf if} (length of intermediate phrase $\geq$ 2) $\wedge$ (accent $=$ deaccented) {\bf then} {\it sentential} \\ 
{\bf if} (length of intermediate phrase $\geq$ 8) $\wedge$ (length of intonational phrase $\leq$ 9) $\wedge$ (accent $=$ L*) {\bf then} {\it sentential} \\ 
default is on {\it discourse} \\
\end{tabbing}
\caption{\label{output-p} Example C4.5 and {\sc cgrendel} classification models learned from different prosodic feature representations of the classifiable cue phrases in the multiple cue phrase corpus.}
\end{center}}
\end{figure}

An examination of some of the best performing learned classification models
shows that they are quite comparable in content to relevant portions
of the prosodic model of Figure~\ref{prosody}, and often contain further
linguistic insights.  Consider the classification model learned from
the single feature {\it position in intonational phrase} (P-P), shown
near the top of Figure~\ref{output-p}.  Both of the learned
classification models say that if the cue phrase is not in the initial
position of the intonational phrase, classify as {\it sentential};
otherwise classify as {\it discourse}.  Note the correspondence with
line (8) in the manually derived prosodic model.  Also note that the
classification models are comparable\footnote{The different feature
values in the two figures reflect the fact that phrasal position was
represented in the ``now'' corpus using symbolic values (as in
Figure~\ref{prosody}), and in the multiple cue phrase corpus using
integers (as in Figure~\ref{features}).} to the P-P classification
models learned from Experiment Set 1 (shown in
Figure~\ref{output-now}), despite the differences in training data.
The fact that the single prosodic feature {\it position in
intonational phrase} (P-P) can classify cue phrases at least as well as the more
complicated manual and multiple feature learned models is again a new result of the learning experiments.

Figure~\ref{output-p} also illustrates the more complex classification
models learned using {\it prosody}, the largest
prosodic feature set.  The C4.5 model is similar to lines (1)
and (8) of the manual model.  (The length value 1 is
equivalent to the composition value {\it alone}.)  
In the ruleset
induced from {\it prosody} by {\sc cgrendel}, the first 2 if-then
rules correlate {\it sentential} status with (among other things)
non-initial position\footnote{Tests such as ``feature $\geq$ x''
and ``feature $\leq$ y'' are merged in the figure for
simplicity, e.g., ``y $\geq$ feature $\geq$ x.''}, and the second 2
rules with H* and H*+L accents; these rules are similar to lines
(6)-(8) in Figure~\ref{prosody}.  However, the last 2 if-then rules in
the ruleset also correlate no accent and L* with sentential status
when the phrase is of a certain length, while lines (4) and (5) in
Figure~\ref{prosody} provide a different interpretation and 
do not take length into account.  Recall that length was coded by Hirschberg
and Litman only in their test data.  Length was thus never used to
generate or revise their prosodic model.  The utility of length is
a new result of this experiment set.

Although not shown, the models learned from {\it phrasing}, {\it
position}, and {\it intonational} also outperform the 
manual model.  As can be seen from
Table~\ref{samples}, these models correspond to all of the 
feature sets that are supersets of P-P but subsets of {\it
prosody}.

\subsubsection{Textual Models}

\begin{table*}[tb]
{\scriptsize
\begin{center}
\begin{tabular}{|l | r| r|} \hline\hline
Model	&Classifiable Cue Phrases (N=878)	&Classifiable Non-Conjuncts (N=495) \\ \hline\hline 
C-P			&(40.7 $\pm$ 6.2)	&(40.2 $\pm$ 4.5)	\\ \hline	 	
C-S			&(41.3 $\pm$ 5.9)	&(39.8 $\pm$ 4.2)	\\ \hline		 	
O-P			&20.6 $\pm$ 5.7		&17.6 $\pm$ 3.3		\\ \hline 		 	
O-P*			&18.4 $\pm$ 3.7		&17.2 $\pm$ 2.4		\\ \hline 		 	
O-S			&(34.1 $\pm$ 6.3)	&(30.2 $\pm$ 1.8)	\\ \hline 		        
O-S*			&(35.2 $\pm$ 5.5)	&(32.6 $\pm$ 3.0)	\\ \hline
POS			&(37.7 $\pm$ 4.1)	&(38.2 $\pm$ 4.6)	\\ \hline \hline
text			&18.8 $\pm$ 4.2		&19.0 $\pm$ 3.6	 	\\ \hline 
adjacency		&(39.7 $\pm$ 5.7)	&(40.2 $\pm$ 3.4)	\\ \hline
orthography		&18.9 $\pm$ 3.4 	&18.8 $\pm$ 3.0		\\ \hline
preceding 		&18.8 $\pm$ 3.8		&17.6 $\pm$ 3.2		\\ \hline 
succeeding		&(33.9 $\pm$ 6.0) 	&(30.0 $\pm$ 2.7)	\\ \hline \hline
manual textual		&19.9 $\pm$ 2.8  	&16.1 $\pm$ 3.4 	\\ \hline
\end{tabular}
\caption{\label{textual} 95\%-confidence intervals for the error rates (\%) of all
{\sc cgrendel} textual classification models, testing data. (Training and testing were done from the multiple cue phrase corpus using cross-validation.) }
\end{center}
}
\end{table*}
Table~\ref{textual} presents the error rates of the classification
models learned by {\sc cgrendel}, in the 24
different textual experiments.  
Unlike the experiments involving the prosodic feature sets, none of
the learned textual models perform significantly better than the
manually derived model.  However, the results suggest that machine
learning is still an effective technique for automating the
development of cue phrase classification models.  In particular, five
learned models (O-P, O-P*, {\it text}, {\it orthography},
and {\it preceding}) perform comparably to the manually derived model,
in both test sets.
Note that these five models are learned from the five textual feature
sets that include either the feature O-P or O-P* (recall
Figure~\ref{features} and Table~\ref{samples}). These models
perform significantly better than all of the remaining learned textual models.

Figure~\ref{output-t} shows the best performing learned textual models.
Note the similarity to the manually
derived model.
\begin{figure}[tb]
{\scriptsize
\begin{center}
\begin{tabbing}
ssss \= ssss \= sssssssssssssssssssssssssssssssssssssssssssssssssssssssssssssssssssssssss\= \kill
\underline{Manually derived textual model (repeated from Figure~\ref{prosody})}:\\ 
{\bf if} preceding orthography $=$ true {\bf then}  {\it discourse} \\
{\bf elseif} preceding orthography $=$ false {\bf then} {\it sentential} \\
\\
\underline{Decision tree learned from O-P*, from text, from orthography, and from preceding using C4.5}:\\ \\
{\bf if} preceding orthography* = NA {\bf then} {\it discourse}\\ {\bf
elseif} preceding orthography* = false {\bf then} {\it sentential}\\
{\bf elseif} preceding orthography* = true {\bf then} {\it
discourse}\\
\\
\underline{Ruleset learned from O-P, from O-P*, from orthography, and from preceding using CGRENDEL}: \\ \\
{\bf if} preceding orthography* = false {\bf then} {\it sentential} \\
default is on {\it discourse} \\
\\
\underline{Ruleset learned from text using CGRENDEL}: \\ \\
{\bf if} preceding orthography* = false {\bf then} {\it sentential} \\
{\bf if} part-of-speech = article {\bf then} {\it sentential} \\
default is on {\it discourse} \\
\end{tabbing}
\caption{\label{output-t} Example C4.5 and {\sc cgrendel} classification models learned from different textual feature representations of the classifiable cue phrases in the multiple cue phrase corpus.}
\end{center}}
\end{figure}
As with the prosodic results, the best performing single feature
models perform comparably to those learned from multiple features.  In
fact, in {\sc cgrendel}, the rulesets learned from the multiple
feature sets {\it orthography} and {\it preceding} are identical to
the rulesets learned from the single features O-P and O-P*, even though more features
were available for use. (The corresponding error rates in
Table~\ref{textual} are not identical due to the estimation using
cross-validation.)  The {\sc cgrendel} model {\it text} also incorporates the 
feature {\it part-of-speech}.
In C4.5, the models {\it text}, {\it orthography}
and {\it preceding} are all identical to O-P*.

\subsubsection{Prosodic/Textual Models}

Table~\ref{combined} presents the error rates of the classification
models learned by {\sc cgrendel} when the data is represented using
{\it speech-text}, the complete set of prosodic and textual features
(recall Table~\ref{samples}).  Since Hirschberg and Litman
did not develop a similar classification model that combined both
types of features, for comparison the last two rows show the error
rates of the separate prosodic and textual models.  Only when the
learned model is compared to the manual prosodic model, using the
classifiable cue phrases for testing, does learning result in a significant
performance improvement.  This is consistent with the results
discussed above, where several learned prosodic models performed
better than the manually derived prosodic model in this test set.
The performance of {\it speech-text} is not significantly better or worse than
the performance of either the best prosodic or textual learned models
(Tables~\ref{prosodic} and~\ref{textual}, respectively).
\begin{table*}[tb]
{\scriptsize
\begin{center}
\begin{tabular}{|l | r| r|} \hline\hline
Model	&Classifiable Cue Phrases (N=878)	&Classifiable Non-Conjuncts (N=495) \\ \hline\hline 
speech-text		&{\it 15.9 $\pm$ 3.2}		&14.6 $\pm$ 4.6		\\ \hline \hline
manual prosodic		&24.6 $\pm$ 3.0		&14.7 $\pm$ 3.2		\\ \hline
manual textual		&19.9 $\pm$ 2.8  	&16.1 $\pm$ 3.4 	\\ \hline
\end{tabular}
\caption{\label{combined} 95\%-confidence intervals for the error rates (\%) of the 
{\sc cgrendel} prosodic/textual classification model, testing data. (Training and testing were done from the multiple cue phrase corpus using cross-validation.) }
\end{center}
}
\end{table*}

\begin{figure}
{\scriptsize
\begin{center}
\begin{tabbing}
ssss \= ssss \= ssss \= ssss \= ssssssssssssssssssssssssssssssssssssssssssssssssssssssssssssssssss\=\=\= \kill
\underline{Manually derived prosodic model (repeated from Figure~\ref{prosody})}:\\ \\
{\bf if} composition of intermediate phrase $=$ alone {\bf then}  {\it discourse} \>\>\>\>\>(1) \\
{\bf elseif}  composition of intermediate phrase $=$ $\neg$alone {\bf then} \>\>\>\>\>(2) \\  
\>{\bf if} position in intermediate phrase $=$ first {\bf then}\>\>\>\>(3)\\
\>	\>{\bf if} accent $=$  deaccented {\bf then}  {\it discourse} \>\>\>(4) \\ 
\>	\>{\bf elseif} accent $=$   L* {\bf then}  {\it discourse}\>\>\>(5) \\
\>	\>{\bf elseif} accent $=$   H* {\bf then}  {\it sentential}\>\>\>(6) \\
\>	\>{\bf elseif} accent $=$  complex {\bf then} {\it sentential}\>\>\>(7) \\
\>{\bf elseif} position in intermediate phrase $=$ $\neg$first {\bf then} {\it sentential}\>\>\>\>(8) \\
\\ 
\underline{Manually derived textual model (repeated from Figure~\ref{prosody})}:\\ 
{\bf if} preceding orthography $=$ true {\bf then}  {\it discourse} \>\>\>\>\>(9)\\
{\bf elseif} preceding orthography $=$ false {\bf then} {\it sentential} \>\>\>\>\>(10)\\
\\
\underline{Decision tree learned from speech-text using C4.5}:\\ \\  
{\bf if} position in intonational phrase $\leq$ 1 {\bf then}\\
\>{\bf if} preceding orthography* = NA {\bf then} {\it discourse} \\
\>{\bf elseif} preceding orthography* = true {\bf then} {\it discourse} \\
\>{\bf elseif} preceding orthography* = false {\bf then}\\
\>\>{\bf if} length of intermediate phrase $>$ 12 {\bf then} {\it discourse} \\
\>\>{\bf elseif} length of intermediate phrase $\leq$ 12 {\bf then}\\
\>\>\>{\bf if} length of intermediate phrase $\leq$ 1 {\bf then} {\it discourse} \\
\>\>\>{\bf elseif} length of intermediate phrase $>$ 1 {\bf then} {\it sentential} \\
{\bf elseif} position in intonational phrase $>$ 1 {\bf then}\\
\>{\bf if} length of intermediate phrase $\leq$ 1 {\bf then} {\it discourse} \\
\>{\bf elseif} length of intermediate phrase $>$ 1 {\bf then} {\it sentential}\\
\\
\underline{Ruleset learned from speech-text using CGRENDEL}:\\ 
\\
{\bf if} (preceding orthography $=$ false) $\wedge$ (4 $\leq$ position in intonational phrase $\leq$ 6) $\wedge$ {\bf then} {\it sentential} \\ 
{\bf if} (preceding orthography $=$ false) $\wedge$ (length of intermediate phrase $\geq$ 2) {\bf then} {\it sentential} \\ 
{\bf if} (preceding orthography $=$ false) $\wedge$ (length of intonational phrase $\geq$ 7) $\wedge$ (preceding cue phrase $=$ NA) \\
\>$\wedge$ (accent $=$ H*) {\bf then} {\it sentential} \\ 
{\bf if} (preceding orthography $=$ comma) $\wedge$ (length of intermediate phrase $\geq$ 5) $\wedge$ (length of intonational phrase $\leq$ 17) \\
\>$\wedge$ (part-of-speech $=$ adverb) {\bf then} {\it sentential} \\ 
{\bf if} (preceding orthography $=$ comma) $\wedge$ (3 $\leq$ length of intonational phrase $\leq$ 8) $\wedge$ (accent $=$ H*) {\bf then} {\it sentential} \\ 
{\bf if} (preceding orthography $=$ comma) $\wedge$ (3 $\leq$ length of intermediate phrase $\leq$ 8) \\
\> $\wedge$ (length of intonational phrase $\geq$ 15) {\bf then} {\it sentential} \\ 
{\bf if} (position in intonational phrase $\geq$ 2) $\wedge$ (length of intermediate phrase $\geq$2) \\
\> $\wedge$ (preceding cue phrase $=$ NA) {\bf then} {\it sentential} \\ 
default is on {\it discourse} \\
\end{tabbing}
\caption{\label{output-c} C4.5 and {\sc cgrendel} classification models learned from the prosodic/textual feature representation of the classifiable cue phrases in the multiple cue phrase corpus.}
\end{center}}
\end{figure}
Figure~\ref{output-c} shows the C4.5 and {\sc cgrendel} hypotheses
learned from {\it speech-text}.  The C4.5 model classifies cue phrases
using the prosodic and textual features that performed best in
isolation ({\it position in intonational phrase} and {\it preceding
orthography*}, as discussed above), in conjunction with the additional
feature {\it length of intermediate phrase} (which also appears in the
model learned from {\it prosody} in Figure~\ref{output-p}).  Like line
(9) in the manually derived textual model, the learned model
associates the presence of preceding orthography with the class {\it
discourse}. Unlike line (10), however, cue phrases not preceded by
orthography may be classified as either {\it discourse} or {\it
sentential}, based on prosodic feature values (which were not
available for use by the textual model).  The branch of the learned
decision tree corresponding to the last three lines is also similar to
lines (1), (2), and (8) of the manually derived prosodic model. (Recall
that a length value of 1 is equivalent to a composition value {\it
alone}.)

The {\sc cgrendel} model uses similar features to those used by C4.5
as well as the prosodic feature {\it accent} (also used in {\it prosody}
in Figure~\ref{output-p}),
and the textual features {\it part-of-speech} (also used in {\it text} in Figure~\ref{output-t}) and  {\it
preceding cue phrase}.
Like C4.5, and unlike line (10) of the
manually derived textual model, the {\sc cgrendel} model classifies
cue phrases lacking preceding orthography as {\it sentential} only in
conjunction with certain other feature values.  Unlike line (9) in the
manual model, the learned model also classifies some cue phrases with
preceding orthography as {\it sentential} (if the orthography is a
comma, and other feature values are present).  Finally, 
the third and fifth learned rules elaborate line (6) with additional prosodic
as well as textual features, while
the first and last learned rules elaborate line (8).

\subsection{Experiment Set 3: Adding the Feature {\it token}}

Each experiment in the third group replicates an experiment from the
second group, with the exception that the data representation now also
includes the lexical feature {\it token} from Figure~\ref{features}.
These experiments investigate how performance changes when
classification models are allowed to treat different cue phrases
differently.  As will be seen, learning from tokenized
feature sets often further improves the performance of the learned classification models.
In addition, the classification models now contain new linguistic information regarding
particular tokens (e.g., ``so'').

\subsubsection{Prosodic Models}

\begin{table*}[tb]
{\scriptsize
\begin{center}
\begin{tabular}{|l | r| r|} \hline\hline
Model	&Classifiable Cue Phrases (N=878)	&Classifiable Non-Conjuncts (N=495) \\ \hline\hline 
P-L+			&21.8 $\pm$ 4.6		&17.4 $\pm$ 2.7			\\ \hline	
P-P+			&{\it 16.7 $\pm$ 2.8}	&14.8 $\pm$ 5.0 	 	\\ \hline	
I-L+			&20.3 $\pm$ 3.4		&16.0 $\pm$ 3.3 	 	\\ \hline	
I-P+			&25.1 $\pm$ 4.1		&17.0 $\pm$ 3.6 	        \\ \hline	
I-C+			&27.0 $\pm$ 3.6		&18.4 $\pm$ 3.4 	 	\\ \hline	
A+			&19.8 $\pm$ 3.2		&12.8 $\pm$ 3.1	        	\\ \hline	
A*+			&18.6 $\pm$ 3.8		&15.4 $\pm$ 2.8	 		\\ \hline \hline	
prosody+		&{\it 16.7 $\pm$ 2.9}	&15.8 $\pm$ 3.1		 	\\ \hline	
hl93features+		&24.0 $\pm$ 4.5	 	&17.4 $\pm$ 4.3 	 	\\ \hline 	
phrasing+ 		&{\it 14.5 $\pm$ 3.3}	&12.6 $\pm$ 3.3 		\\ \hline 
length+			&{\it 18.6 $\pm$ 2.0}	&16.2 $\pm$ 3.5		 	\\ \hline	 
position+		&{\it 15.6 $\pm$ 3.3}	&13.0 $\pm$ 3.9 	 	\\ \hline	 
intonational+		&{\it 15.1 $\pm$ 2.2}	&16.6 $\pm$ 4.6 	 	\\ \hline	 
intermediate+		&18.5 $\pm$ 3.7		&16.6 $\pm$ 4.0 	 	\\ \hline \hline
manual prosodic		&24.6 $\pm$ 3.0		&14.7 $\pm$ 3.2			\\ \hline
\end{tabular}
\caption{\label{prosodic+} 95\%-confidence intervals for the error rates (\%) of all
{\sc cgrendel} prosodic, {\it tokenized} classification models, testing data. (Training and testing were done from the multiple cue phrase corpus using cross-validation.) }
\end{center}
}
\end{table*}
Table~\ref{prosodic+} presents the error of the learned classification
models on both test sets from the multiple cue phrase corpus, for each
of the {\it tokenized} prosodic feature sets.  Again, the error rates
in italics indicate that the performance of the learned classification
model meaningfully exceeds the performance of the ``manual prosodic''
model (which did not consider the feature {\it
token}).

One way that the improvement obtained by adding the feature {\it
token} can be seen is by comparing the performance of the learned and
manually derived models.  In Table~\ref{prosodic+}, six {\sc cgrendel}
classification models have lower (italicized) error rates than the
manual model. In Table~\ref{prosodic}, only five of these models
are italicized. Thus, adding the feature {\it token} results in an
additional learned model - {\it length+} - outperforming the manually
derived model.
Conversely, in Table~\ref{prosodic+}, no learned models perform
significantly worse than the manually derived manual.  In contrast, in
Table~\ref{prosodic}, several non-tokenized models perform worse than
the manual model (I-C in the larger test set, and 
P-L, I-L, I-C, A, A*, and {\it length}
in the non-conjunct test set).  

The improvement obtained by adding the feature {\it token} can also be
seen by comparing the performance of the tokenized (Table~\ref{prosodic+})
and non-tokenized (Table~\ref{prosodic})
versions of each model to each other.  For convenience, cases where
tokenization yields improvement are highlighted in Table~\ref{p-p+}.
The table shows that the error rate of the tokenized versions of the
feature sets is significantly lower than the error of the
non-tokenized versions, for P-L, I-C, A, A*,
and {\it length} in both test sets, and for I-L in only the
non-conjunct test set. Note the overlap between the feature sets of
Table~\ref{p-p+} and those discussed in the previous paragraph.
\begin{table*}[tb]
{\scriptsize
\begin{center}
\begin{tabular}{|l || r| r||r|r|} \hline\hline 
Model	&\multicolumn{2}{ c||} {Classifiable Cue Phrases (N=878)} &\multicolumn{2}{c|} {Classifiable Non-Conjuncts (N=495)} \\ \hline\hline 
   	&Non-Tokenized &Tokenized (+) 		&Non-Tokenized	&Tokenized (+)  \\ \hline \hline	
P-L	&33.0 $\pm$ 5.9	&21.8 $\pm$ 4.6		&33.2 $\pm$ 1.9	&17.4 $\pm$ 2.7	\\ \hline	
I-L	&-		&	-		&25.6 $\pm$ 2.8	&16.0 $\pm$ 3.3 \\ \hline	
I-C	&36.5 $\pm$ 5.4	&27.0 $\pm$ 3.6		&35.2 $\pm$ 3.4	&18.4 $\pm$ 3.4 \\ \hline	
A	&28.6 $\pm$ 3.6	&19.8 $\pm$ 3.2		&30.2 $\pm$ 3.1	&12.8 $\pm$ 3.1	\\ \hline	
A*	&28.3 $\pm$ 4.3	&18.6 $\pm$ 3.8		&28.4 $\pm$ 1.7	&15.4 $\pm$ 2.8	\\ \hline 
length	&26.1 $\pm$ 3.8	&18.6 $\pm$ 2.0		&27.4 $\pm$ 3.4 &16.2 $\pm$ 3.5 \\ \hline	 
\end{tabular}
\caption{\label{p-p+} Cases where adding the feature {\it token} improves the performance of a prosodic model.}
\end{center}
}
\end{table*}

Figure~\ref{output-p+}
shows several tokenized single feature prosodic classification models.
\begin{figure}
{\scriptsize
\begin{center}
\begin{tabbing}
ssss \= ssss \= ssss \= ssss \= ssssssssssssssssssssssssssssssssssssssssssssssssssssssssssssssssss\=\=\= \kill
\underline{Manually derived prosodic model (repeated from Figure~\ref{prosody})}:\\ \\
{\bf if} composition of intermediate phrase $=$ alone {\bf then}  {\it discourse} \>\>\>\>\>(1) \\
{\bf elseif}  composition of intermediate phrase $=$ $\neg$alone {\bf then} \>\>\>\>\>(2) \\  
\>{\bf if} position in intermediate phrase $=$ first {\bf then}\>\>\>\>(3)\\
\>	\>{\bf if} accent $=$  deaccented {\bf then}  {\it discourse} \>\>\>(4) \\ 
\>	\>{\bf elseif} accent $=$   L* {\bf then}  {\it discourse}\>\>\>(5) \\
\>	\>{\bf elseif} accent $=$   H* {\bf then}  {\it sentential}\>\>\>(6) \\
\>	\>{\bf elseif} accent $=$  complex {\bf then} {\it sentential}\>\>\>(7) \\
\>{\bf elseif} position in intermediate phrase $=$ $\neg$first {\bf then} {\it sentential}\>\>\>\>(8) \\
\\ 
\underline{Ruleset learned from P-L+ using CGRENDEL}:\\ \\
{\bf if} length of intonational phrase $\leq$ 1 {\bf then} {\it discourse} \\
{\bf if} (7 $\leq$ length of intonational phrase $\leq$ 11) $\wedge$ (token $=$ although) {\bf then} {\it discourse} \\
{\bf if} (9 $\leq$ length of intonational phrase $\leq$ 16) $\wedge$ (token $=$ indeed) {\bf then} {\it discourse} \\
{\bf if} (length of intonational phrase $\leq$ 20) $\wedge$ (token $=$ say) {\bf then} {\it discourse} \\
{\bf if} (11 $\leq$ length of intonational phrase $\leq$ 13) $\wedge$ (token $=$ then) {\bf then} {\it discourse} \\
{\bf if} (length of intonational phrase $=$ 5) $\wedge$ (token $=$ well) {\bf then} {\it discourse} \\
{\bf if} token $=$ finally {\bf then} {\it discourse}  \\
{\bf if} token $=$ further {\bf then} {\it discourse}  \\
{\bf if} token $=$ however {\bf then} {\it discourse}  \\
{\bf if} token $=$ now {\bf then} {\it discourse}  \\
{\bf if} token $=$ ok {\bf then} {\it discourse}  \\
{\bf if} token $=$ otherwise {\bf then} {\it discourse}  \\
{\bf if} token $=$ so {\bf then} {\it discourse}  \\
default is on {\it sentential}\\
\\ 
\underline{Ruleset learned from I-C+ using CGRENDEL}:\\ \\
{\bf if} composition of intermediate phrase $=$ only {\bf then} {\it discourse} \\
{\bf if} token $=$ finally {\bf then} {\it discourse}  \\
{\bf if} token $=$ however {\bf then} {\it discourse}  \\
{\bf if} token $=$ now {\bf then} {\it discourse}  \\
{\bf if} token $=$ ok {\bf then} {\it discourse}  \\
{\bf if} token $=$ say {\bf then} {\it discourse}  \\
{\bf if} token $=$ so {\bf then} {\it discourse}  \\
default is on {\it sentential}\\
\\
\underline{Ruleset learned from A+ using CGRENDEL}:\\ \\
{\bf if} accent $=$ L* {\bf then} {\it discourse} \\
{\bf if} (accent $=$ deaccented) $\wedge$ (token $=$ say) {\bf then} {\it discourse} \\
{\bf if} (accent $=$ deaccented) $\wedge$ (token $=$ so) {\bf then} {\it discourse}  \\
{\bf if} (accent $=$  L+H*) $\wedge$ (token $=$ further) {\bf then} {\it discourse}  \\
{\bf if} (accent $=$  L+H*) $\wedge$ (token $=$ indeed) {\bf then} {\it discourse} \\
{\bf if} token $=$ finally {\bf then} {\it discourse}  \\ 
{\bf if} token $=$ however {\bf then} {\it discourse}  \\
{\bf if} token $=$ now {\bf then} {\it discourse}  \\
{\bf if} token $=$ ok {\bf then} {\it discourse}  \\
default is on {\it sentential}\\
\end{tabbing}
\vspace{-.2in}
\caption{\label{output-p+} Example {\sc cgrendel} classification models learned from different {\it tokenized}, prosodic feature representations of the classifiable non-conjuncts in the multiple cue phrase corpus.}
\end{center}}
\end{figure}
The first {\sc cgrendel} model in the figure shows the 
ruleset learned from P-L+, which reduces the 33.2\% $\pm$ 1.9\% error rate of P-L
({\it length of intonational phrase}) to 17.4\% $\pm$ 2.7\%, when
trained and tested using the classifiable non-conjuncts (Table~\ref{p-p+}).  Note that
the first rule uses only a prosodic feature (like the rules of
Experiment Sets 1 and 2), and is in fact similar to line (1) of the manual model.
(Recall that the length value 1 is
equivalent to the composition value {\it alone}.)  
However, unlike the rules of the previous
experiment sets, the next 5 rules use both the prosodic feature and the
lexical feature {\it token}. Also unlike the rules of the previous
experiment sets, the remaining rules classify cue phrases using only
the feature {\it token}.  Examination of the learned rulesets in
Figures~\ref{output-p+} and~\ref{output-t+}
shows that the same cue phrases often appear in this last type of rule.
Some of these cue phrases, for example, ``finally'', ``however'', and
``ok'', are in fact always {\it discourse} usages in the multiple cue
phrase corpus.  For the other cue phrases, classifying cue phrases
using only {\it token} corresponds to classifying cue phrases using
their default class (the most frequent type of usage in the multiple
cue phrase corpus).  Recall the use of a non-tokenized default class
model in Table~\ref{hl93error}.

The second example shows the ruleset learned from I-C+ ({\it
composition of intermediate phrase+}). The first rule corresponds to
line (1) of the manually derived model.\footnote{As discussed in relation to Figure~\ref{features},
the I-C values {\it only} and {\it only cue phrases} in the
multiple cue phrase corpus replace the value {\it alone} in the
``now'' corpus.} The next six rules classify particular cue phrases as
{\it discourse}, independently of the value of I-C.  Note that
although in this model the cue phrase ``say'' is classified using only
{\it token}, in the previous model a more sophisticated strategy for
classifying ``say'' could be found.

The third example shows the {\sc cgrendel} ruleset learned from A+ ({\it accent+}).
The
first rule corresponds to line (5) of the manually derived prosodic
model. In contrast to line (4), however, {\sc cgrendel} uses
deaccenting to predict {\it discourse} for only the tokens ``say'' and
``so.'' If the token is ``finally'', ``however'', ``now'' or ``ok'',
{\it discourse} is assigned (for all accents).  In all other
deaccented cases, {\it sentential} is assigned (using the default).
Similarly, in contrast to line (7), the complex accent L+H* predicts
{\it discourse} for the cue phrases ``further'' and ``indeed'' (and
also for ``finally'', ``however'', ``now'' and ``ok''), and {\it
sentential} otherwise.  

To summarize, new prosodic results of Experiment Set 3 are that features
relating to length, composition, and accent, while not
useful (in isolation) for predicting the classification of all cue
phrases, are in fact quite useful for predicting the class of
individual cue phrases or subsets of cue phrases.  
(Recall that the result of Experiment Sets 1 and 2 was that without
{\it token}, only the prosodic feature {\it position in intonational phrase}
was useful in isolation.)

\subsubsection{Textual Models}

\begin{table*}[tb]
{\scriptsize
\begin{center}
\begin{tabular}{|l | r| r|} \hline\hline
Model	&Classifiable Cue Phrases (N=878)	&Classifiable Non-Conjuncts (N=495) \\ \hline\hline 
C-P+			&(28.2 $\pm$ 3.9)	&16.4 $\pm$ 4.6		\\ \hline	 	
C-S+			&(28.9 $\pm$ 3.6)	&17.2 $\pm$ 4.0		\\ \hline		 	
O-P+			&17.5 $\pm$ 4.4		&10.0 $\pm$ 3.1		\\ \hline 		 	
O-P*+			&17.7 $\pm$ 2.9		&12.2 $\pm$ 2.9		\\ \hline 		 	
O-S+			&26.9 $\pm$ 4.7		&18.4 $\pm$ 3.9		\\ \hline 		        
O-S*+			&(27.3 $\pm$ 3.5)	&16.0 $\pm$ 3.2		\\ \hline
POS+			&(27.4 $\pm$ 3.6)	&17.2 $\pm$ 3.9		\\ \hline \hline
text+			&18.4 $\pm$ 3.0		&12.0 $\pm$ 2.6	 	\\ \hline 
adjacency+		&(28.6 $\pm$ 4.1)	&15.2 $\pm$ 3.1		\\ \hline
orthography+		&17.6 $\pm$ 3.0 	&13.6 $\pm$ 3.9		\\ \hline
preceding+ 		&17.0 $\pm$ 4.1		&13.6 $\pm$ 2.6		\\ \hline 
succeeding+		&25.6 $\pm$ 3.9 	&18.0 $\pm$ 4.5		\\ \hline \hline
manual textual		&19.9 $\pm$ 2.8  	&16.1 $\pm$ 3.4 	\\ \hline
\end{tabular}
\caption{\label{textual+} 95\%-confidence intervals for the error rates (\%) of all
{\sc cgrendel} textual, {\it tokenized} classification models, testing data. (Training and testing were done from the multiple cue phrase corpus using cross-validation.) }
\end{center}
}
\end{table*}
Table~\ref{textual+} presents the error of the learned classification
models on both test sets from the multiple cue phrase corpus, for each
of the {\it tokenized} textual feature sets.  As in Experiment Set 2
(Table~\ref{textual}), none of the {\sc cgrendel} classification
models have lower (italicized) error rates than the manual model.
However, adding the feature {\it token} does improve the performance
of many of the learned rulesets, in that the following models (unlike
their non-tokenized counterparts) are no longer outperformed by the
manual model: O-S+ and {\it succeeding+} in the larger test set, and 
C-P+, C-S+, O-S+, O-S*+, POS+, {\it adjacency+}, and {\it succeeding+} in the non-conjunct test set.

The improvement obtained by adding the feature {\it token} can also be
seen by comparing the performance of the tokenized (Table~\ref{textual+})
and non-tokenized (Table~\ref{textual})
versions of each model to each other, as shown in Table~\ref{t-t+}.
The table shows that the error rates of the tokenized versions of the
feature sets are significantly lower than the error of the
non-tokenized versions, for C-P, C-S, POS, and {\it
adjacency} in both test sets, and for O-P, O-S, 
O-S*, {\it text}, and {\it succeeding} in the non-conjunct test set.
Note the overlap between the feature sets of Table~\ref{t-t+} and
those discussed in the previous paragraph.
\begin{table*}[tb]
{\scriptsize
\begin{center}
\begin{tabular}{|l || r| r||r|r|} \hline\hline 
Model	&\multicolumn{2}{ c||} {Classifiable Cue Phrases (N=878)} &\multicolumn{2}{c|} {Classifiable Non-Conjuncts (N=495)} \\ \hline\hline 
   	&Non-Tokenized	&Tokenized (+) 		&Non-Tokenized	&Tokenized (+) 		\\ \hline \hline
C-P	&40.7 $\pm$ 6.2	&28.2 $\pm$ 3.9		&40.2 $\pm$ 4.5	&16.4 $\pm$ 4.6		\\ \hline	
C-S	&41.3 $\pm$ 5.9	&28.9 $\pm$ 3.6		&39.8 $\pm$ 4.2	&17.2 $\pm$ 4.0		\\ \hline	
O-P	&-		&-			&17.6 $\pm$ 3.3	&10.0 $\pm$ 3.1		\\ \hline  	
O-S	&-		&-			&30.2 $\pm$ 1.8	&18.4 $\pm$ 3.9		\\ \hline	  
O-S*	&-		&-			&32.6 $\pm$ 3.0	&16.0 $\pm$ 3.2		\\ \hline
POS	&37.7 $\pm$ 4.1	&27.4 $\pm$ 3.6		&38.2 $\pm$ 4.6	&17.2 $\pm$ 3.9		\\ \hline 
text	&-		&-			&19.0 $\pm$ 3.6	&12.0 $\pm$ 2.6	 	\\ \hline 
adjacency&39.7 $\pm$ 5.7&28.6 $\pm$ 4.1		&40.2 $\pm$ 3.4	&15.2 $\pm$ 3.1		\\ \hline
succeeding&-		&-		 	&30.0 $\pm$ 2.7	&18.0 $\pm$ 4.5		\\ \hline
\end{tabular}
\caption{\label{t-t+} Cases where adding the feature {\it token} improves the performance of a textual model.}
\end{center}
}
\end{table*}

Figure~\ref{output-t+} shows
several tokenized single textual feature classification models.
\begin{figure}
{\scriptsize
\begin{center}
\begin{tabbing}
\underline{Manually derived textual model (repeated from Figure~\ref{prosody})}:\\ \\
{\bf if} preceding orthography $=$ true {\bf then}  {\it discourse}\\
{\bf elseif} preceding orthography $=$ false {\bf then} {\it sentential} \\
\\
\underline{Ruleset learned from C-P+ using CGRENDEL}:\\ \\
{\bf if} (preceding cue phrase $=$ true) $\wedge$ (token $=$ indeed) {\bf then} {\it discourse} \\
{\bf if} (preceding cue phrase $=$ NA) $\wedge$ (token $=$ further) {\bf then} {\it discourse} \\
{\bf if} (preceding cue phrase $=$ NA) $\wedge$ (token $=$ now) {\bf then} {\it discourse} \\
{\bf if} (preceding cue phrase $=$ NA) $\wedge$ (token $=$ so) {\bf then} {\it discourse} \\
{\bf if} token $=$ although {\bf then} {\it discourse}  \\
{\bf if} token $=$ finally {\bf then} {\it discourse}  \\ 
{\bf if} token $=$ however {\bf then} {\it discourse}  \\
{\bf if} token $=$ ok {\bf then} {\it discourse}  \\
{\bf if} token $=$ say {\bf then} {\it discourse}  \\
{\bf if} token $=$ similarly {\bf then} {\it discourse}  \\
default is on {\it sentential} \\
\\
\underline{Ruleset learned from O-P+ using CGRENDEL}:\\ \\
{\bf if} preceding orthography $=$ false {\bf then} {\it sentential} \\
{\bf if} (preceding orthography $=$ comma) $\wedge$ (token $=$ then) {\bf then} {\it sentential} \\
default is on {\it discourse}\\
\\
\underline{Ruleset learned from O-S+ using CGRENDEL}:\\ \\
{\bf if} succeeding orthography $=$ comma {\bf then} {\it discourse} \\
{\bf if} (succeeding orthography $=$ false) $\wedge$ (token $=$ so) {\bf then} {\it discourse} \\
{\bf if} succeeding orthography $=$ NA {\bf then} {\it discourse} \\
{\bf if} token $=$ although {\bf then} {\it discourse}  \\
{\bf if} token $=$ finally {\bf then} {\it discourse}  \\ 
{\bf if} token $=$ now {\bf then} {\it discourse}  \\ 
{\bf if} token $=$ ok {\bf then} {\it discourse}  \\
{\bf if} token $=$ say {\bf then} {\it discourse}  \\
default is on {\it sentential} \\ 
\\
\underline{Ruleset learned from POS+ using CGRENDEL}:\\ \\
{\bf if} (part-of-speech $=$ adverb) $\wedge$ (token $=$ finally) {\bf then} {\it discourse} \\
{\bf if} (part-of-speech $=$ singular proper noun) $\wedge$ (token $=$ further) {\bf then} {\it discourse} \\
{\bf if} (part-of-speech $=$ adverb) $\wedge$ (token $=$ however) {\bf then} {\it discourse} \\
{\bf if} (part-of-speech $=$ adverb) $\wedge$ (token $=$ indeed) {\bf then} {\it discourse} \\
{\bf if} (part-of-speech $=$ subordinating conjunction) $\wedge$ (token $=$ so) {\bf then} {\it discourse} \\
{\bf if} token $=$ although {\bf then} {\it discourse}  \\
{\bf if} token $=$ now {\bf then} {\it discourse}  \\
{\bf if} token $=$ say {\bf then} {\it discourse}  \\
{\bf if} token $=$ ok {\bf then} {\it discourse}  \\
default is on {\it sentential} \\
\end{tabbing}
\caption{\label{output-t+} Example {\sc cgrendel} classification models learned from different {\it tokenized}, textual feature representations of the classifiable non-conjuncts in the multiple cue phrase corpus.}
\end{center}}
\end{figure}
The first {\sc cgrendel} model shows the ruleset learned from C-P+ ({\it preceding
cue phrase+}),
which reduces the 40.2\% $\pm$ 4.5\% error rate of C-P to 16.4\% $\pm$ 4.6\% when trained and
tested using the classifiable non-conjuncts (Table~\ref{t-t+}).  This
ruleset correlates preceding cue phrases with
discourse usages of ``indeed'',  and omitted 
transcriptions of ``further'', ``now'', and ``so'' 
with discourse usages.
The classifications for the rest of
the cue phrases are predicted using only the feature token.

The second example shows the {\sc cgrendel} ruleset learned from O-P+ ({\it preceding
orthography+}).
This ruleset correlates no preceding orthography with sentential usages of
cue phrases (as in both the manually derived model 
and the learned models from Experiment Set 2).  Unlike
those models, however, the cue phrase ``then'' is also classified as
{\it sentential}, even when it is preceded by orthography (namely,
by a comma).  

The third example shows the {\sc cgrendel}
ruleset learned from O-S+ ({\it succeeding
orthography}).  This ruleset correlates the presence of succeeding
commas with discourse usages of cue phrases, except for the cue
phrase ``so'', which is classified as a discourse usage without any succeeding
orthography. The model also correlates cue phrases that were omitted from the
transcript with discourse usages.  The classifications for the rest of
the cue phrases are predicted using only the feature {\it token}.

The last example shows the {\sc cgrendel} ruleset learned
from POS+ ({\it part-of-speech+}).  This ruleset classifies certain cue phrases
as discourse usages depending on both {\it part-of-speech} and {\it token}, as well
as independently of {\it part-of-speech}.

\begin{figure}
{\scriptsize
\begin{center}
\begin{tabbing}
ssss \= ssss \= sssssssssssssssssssssssssssssssssssssssssssssssssssssssssssssssssssssssss\= \kill
\underline{Ruleset learned from text+ using CGRENDEL}: \\ \\
{\bf if} preceding orthography $=$ false {\bf then} {\it sentential} \\
{\bf if} (preceding orthography $=$ comma) $\wedge$ (token $=$ although) {\bf then} {\it sentential} \\
{\bf if} (preceding orthography $=$ comma) $\wedge$ (token $=$ no) {\bf then} {\it sentential} \\
{\bf if} (preceding orthography $=$ comma) $\wedge$ (token $=$ then) {\bf then} {\it sentential} \\
{\bf if} (succeeding orthography $=$ false) $\wedge$ (preceding cue phrase $=$ NA) $\wedge$ (token $=$ similarly) {\bf then} {\it sentential} \\
{\bf if} token $=$ actually {\bf then} {\it sentential} \\
{\bf if} token $=$ first {\bf then} {\it sentential} \\
{\bf if} token $=$ since {\bf then} {\it sentential} \\
{\bf if} token $=$ yes {\bf then} {\it sentential} \\
default is on {\it discourse} \\
\end{tabbing}
\caption{\label{output-text} {\sc cgrendel} classification model learned from
a tokenized, {\it multiple} textual feature representation of the classifiable non-conjuncts in the multiple cue phrase corpus.}
\end{center}}
\end{figure}
Finally, Figure~\ref{output-text} shows the classification
model learned from {\it text+}, the largest tokenized
textual feature set.  Note that three of the four features used in the
tokenized, single textual feature models of Figure~\ref{output-t+} are
incorporated into this tokenized, multiple textual feature model.

To summarize, new textual results of Experiment Set 3 are that 
features based on adjacent cue phrases, succeeding orthography, and
part-of-speech, while not useful (in isolation) for predicting the
classification of all cue phrases, are in fact quite useful in
conjunction with only the feature {\it token}.  (Recall that the
result of Experiment Set 2 was that without {\it token}, only the
textual features {\it preceding orthography} and {\it preceding
orthography*} were useful in isolation.)

\subsubsection{Prosodic/Textual Models}

\begin{table*}[tb]
{\scriptsize
\begin{center}
\begin{tabular}{|l | r| r|} \hline\hline
Model	&Classifiable Cue Phrases (N=878)	&Classifiable Non-Conjuncts (N=495) \\ \hline\hline 
speech-text+		&{\it 16.9 $\pm$ 3.4}	&16.6 $\pm$ 4.1		\\ \hline \hline
manual prosodic		&24.6 $\pm$ 3.0		&14.7 $\pm$ 3.2		\\ \hline
manual textual		&19.9 $\pm$ 2.8  	&16.1 $\pm$ 3.4 	\\ \hline
\end{tabular}
\caption{\label{combined+} 95\%-confidence intervals for the error rates (\%) of the
{\sc cgrendel} prosodic/textual, {\it tokenized} classification models, testing data. (Training and testing were done from the multiple cue phrase corpus using cross-validation.) }
\end{center}
}
\end{table*}
Table~\ref{combined+} presents the error rates of the classification
models learned by {\sc cgrendel} when the data is represented using
{\it speech-text+}, the complete set of prosodic and textual features.
As in Experiment Set 2, the
performance of {\it speech-text+} is not better than the performance
of either the best learned (tokenized) prosodic or textual models
(Tables~\ref{prosodic+} and~\ref{textual+}, respectively).

Comparison of Tables~\ref{combined} and~\ref{combined+} also shows
that for the feature set {\it speech-text}, tokenization does not improve
performance.  This is in contrast to the prosodic and textual feature
sets, where tokenization improves the performance of many learned
models (namely those shown in Tables~\ref{p-p+} and~\ref{t-t+}).

\subsection{Experiment Set 4: Adding the Classification {\it ambiguous}}
\label{exptfour}

In practice, a cue phrase classification model will have to classify
{\it all} the cue phrases in a recording or text, not just those that
are ``classifiable.''  The experiments in the fourth set replicate the
experiments in Experiment Sets 1, 2, and 3, with the exception that
all 953 cue phrases in the multiple cue phrase corpus are now used.  This
means that cue phrases are now classified as {\it
discourse}, {\it sentential}, as well as {\it unknown} (defined in
Table~\ref{judge}).  Experiment Set 4 investigates whether machine
learning can explicitly recognize the new class {\it unknown}.

Recall that the studies of Hirschberg and Litman did not attempt to
predict the class {\it unknown}, as it did not occur in their ``now''
training corpus.  Thus in Experiment Set 1, the class {\it unknown}
similarly can not be learned from the training data.  However, the
{\it unknown} examples can be added to the testing data of Experiment
Set 1.  Obviously performance will degrade, as the models must
incorrectly classify each {\it unknown} example as either {\it
discourse} or {\it sentential}.  For example, when tested on the full
corpus of 953 example cue phrases, the 95\% confidence intervals for
the error rates of P-P and {\it intonational} are 24.8\% $\pm$ 2.8\%;
recall that when tested on the subset of the corpus corresponding to
the 878 classifiable cue phrases, the error was 18.3\% $\pm$ 2.6\%
(Table~\ref{nowtesting}).

Unfortunately, the results of rerunning Experiment Sets 2 and 3 
do not show promising results for classifying cue phrases
as {\it unknown}.  Despite the presence of 75 examples of {\it
unknown}, most of the learned models still classify {\it unknown} cue
phrases as only {\it discourse} or {\it sentential}.  For example,
when {\sc cgrendel} is used for learning, only 2 of the possible 27
non-tokenized models\footnote{Recall that Experiment Sets 2 and 3 constructed
14 prosodic models, 12 textual models, and 1 prosodic/textual model.}
({\it phrasing} and {\it speech-text}) contain rules that predict the
class {\it unknown}.  Furthermore, each of these models only contains
one rule for {\it unknown}, and each of these rules only applies to 2
of the possible 953 examples!  Similarly, only four of the possible 27
tokenized models ({\it length+}, {\it phrasing+}, {\it prosody+}, and
{\it speech-text+}) contain at least one rule for the class {\it
unknown}.  When compared to training and testing using only the
classifiable cue phrases in the corpus, the error rate on the full corpus
is typically (but not always) significantly higher.
The best performing model in Experiment Set 4 is {\it speech-text+},
with a 22.4\% $\pm$ 4.1\% error rate (95\% confidence interval).  

In sum, Experiment Set 4 addressed a problem that was previously
unexplored in the literature - the ability to develop classification
models that predict not only {\it discourse} and {\it sentential}
usages of cue phrases, but also usages which human judges find
difficult to classify.  Unfortunately, the results of the experiments
suggest that learning how to classify cue phrases as {\it unknown} is
a difficult problem.  Perhaps with more training data (recall that
there are only 75 examples of {\it unknown}) or with additional
features better results could be obtained.

\subsection{Discussion}

The experimental results suggest that machine learning is a useful
tool for both automating the generation of classification models and
improving upon manually derived results.  In Experiment Sets 1 and 2
the performance of many of the learned classification models is
comparable to the performance of the manually derived models.  In
addition, when tested on the classifiable cue phrases, several learned
prosodic classification models (as well as the learned
prosodic/textual model) outperform Hirschberg and Litman's manually
derived prosodic model.  Experiment Set 3 shows that learning from
tokenized feature sets even further improves performance, especially
in the non-conjunct test set.  More
tokenized than non-tokenized learned models perform at least as well
as the manually derived models.  Many tokenized learned models also
outperform their non-tokenized counterparts. 

While the textual classification models do not outperform the better
prosodic classification models, they have the advantage that the
textual feature values are obtained directly from the transcript,
while determining the values of prosodic features requires manual
analysis. (See, however, Section~\ref{utility} for a discussion of the
feasibility of automating the prosodic analysis. In addition, a
transcript may not always be available.)  On the other hand,
almost all the high performing textual models are dependent on orthography.
While manual transcriptions of prosodic features have been shown to be
reliable across coders~\cite{tobi}, there are no corresponding
results for the reliability of orthography.  

Examination of the best performing learned models shows that they are
often comparable in content to the relevant portions of the manually
derived models.  Examination of the models also provides new
contributions to the cue phrase literature.  For example, Experiment
Sets 1 and 2 demonstrate the utility of classifying cue phrases based
on only a single prosodic feature - phrasal position.\footnote{The
empirical studies performed by \shortciteA{holte:mlj93} show that for many
other datasets, the accuracy of single feature rules and decision
trees is often competitive with the accuracy of more complex learned
models.} Experiment Set 2 also demonstrates the utility of the
prosodic feature {\it length} and the textual feature {\it preceding
cue phrase} for classifying cue phrases - in conjunction with other
prosodic and textual features.  Finally, the results of Experiment Set
3 demonstrate that even though many features are not useful by themselves for
classifying all cue phrases, they may nonetheless be very informative
in their {\it tokenized} form.  This is
true for the prosodic features based on phrasal length, phrasal
composition, and accent, and for the textual features based on
adjacent cue phrases, succeeding position, and part-of-speech.\footnote{In contrast, the prosodic features phrasal composition and accent were
previously known to be useful {\it in conjunction with} each other and with
phrasal position~\cite{hl93}, while part-of-speech was known to be
useful only in conjunction with orthography~\cite{hl93}.
Length, adjacent cue phrases, and succeeding position
were not used in either of the manually derived models~\cite{hl93} (although length
and adjacent cue phrases were shown to be useful - again only in conjunction
with other prosodic and textual features - in Experiment Set 2).}
\section{Utility}
\label{utility}

The results of the machine learning experiments are quite promising,
in that when compared to manually derived classification models
already in the literature, the learned classification models often
perform with comparable if not higher accuracy.  Thus, machine
learning appears to be an effective technique for automating the
generation of classification models.  However, given that the
experiments reported here still rely on manually created training
data, a discussion of the practical utility of the results is in
order.

Even given manually created training data, the results established by
\shortciteA{hl93} - obtained using even less
automation than the experiments of this paper - are already having
practical import.  In particular, the manually derived cue phrase
classification models are used to improve the naturalness of the
synthetic speech in a text-to-speech system~\cite{hirschberg90}.
Using the text-based model, the text-to-speech system classifies each
cue phrase in a text to be synthesized as either a discourse or
sentential usage.  Using the prosodic model, the system then conveys
this usage by synthesizing the cue phrase with the appropriate type of
intonation.  The speech synthesis could be further improved (and the
output made more varied) by using any one of the higher
performing learned prosodic models presented in this paper.

The results of this paper could also be directly applied in the area
of text generation.  For example, 
\shortciteA{Moser95} are concerned with the implementation of
cue selection and placement strategies in natural language generation
systems.
Such systems could be enhanced by
using the text-based models of cue phrase classification (particularly
the tokenized models) to additionally specify preceding and succeeding
orthography, part-of-speech, and adjacent cue phrases that are
appropriate for discourse usages.

Finally, if the results of this paper could be fully automated, they
could also be used in natural language {\it understanding} systems, by
enhancing their ability to recognize discourse structure.  The results
obtained by \shortciteA{LitmanPassonneau95} and \shortciteA{PassonneauLitman97} suggest that
algorithms that use cue phrases (in conjunction with other features)
to predict discourse structure outperform algorithms that do not take
cue phrases into account.  In particular, Litman and Passonneau
develop several algorithms that explore how features of cue phrases,
prosody and referential noun phrases can be best combined to predict
discourse structure.  Quantitative evaluations of their results show
that the best performing algorithms all incorporate the use of  discourse usages of cue
phrases (where cue phrases are
classified as discourse using only phrasal position). As discussed in
Section~\ref{introduction}, discourse structure is useful for
performing tasks such as anaphora resolution and plan recognition.
Recent work has also shown that if discourse
structure can be recognized, it can be used to improve retrieval
of text~\cite{hearst94} and speech~\cite{stifleman95}.

Although the prosodic features were manually labeled by
Hirschberg and Litman, there are recent results suggesting that at
least some aspects of prosody can be automatically labeled directly
from speech.  For example, 
\shortciteA{Wightman94} develop an
algorithm that is able to automatically recognize prosodic phrasing
with 85-86\% accuracy (measured by
comparing automatically derived labels with hand-marked labels); this
accuracy is only slightly less than human-human accuracy. 
Recall
that the experimental results of this paper show that models learned
from the single feature {\it position in intonational phrase} - which
could be automatically computed given such an automatic prosodic
phrasing algorithm - perform at least as well as any other learned
prosodic model.  Similarly, accenting versus deaccenting can be
automatically labeled with 88\% accuracy~\cite{Wightman94}, while a
more sophisticated labeling scheme that distinguishes between four
types of accent classes (and is somewhat similar to the prosodic
feature {\it accent*} used in this paper) can be labeled with 85\%
accuracy~\cite{Ostendorf}.
Recall from Experiment Set 3 that the tokenized models learned using
{\it accent*} also classify cue phrases with good results.

Although the textual features were automatically extracted
from a transcript, the transcript itself was manually created.  Many
natural language understanding systems do not deal with speech at all,
and thus begin with such textual representations.  In spoken language
systems the transcription process is typically automated using a
speech recognition system (although this
introduces further sources of error).

\section{Related Work}
\label{related}

This paper has both compared the results obtained using machine learning
to previously existing manually-obtained results, and has
also used machine learning as a tool for developing theories given new linguistic
data (as in the models resulting from Experiment Set 3, where the new feature {\it token} was considered).
\shortciteA{Siegel93} similarly uses machine learning (in particular, a genetic learning algorithm) to classify cue phrases
from a previously unstudied set of textual features:
a feature corresponding to {\it
token}, as well as textual features containing the lexical or
orthographic item immediately to the left of and in the 4 positions to
the right of the example.  
Siegel's input consists of one judge's non-ambiguous examples taken from the data used by \shortciteA{hl93} as well as additional examples; his output
is in the form of decision trees.
Siegel reports a 21\%
estimated error rate, with half of the corpus used for training and
half for testing.  
\shortciteA{SiegelMcKeown} 
also propose a method for developing linguistically viable rulesets,
based on the partitioning of the training data produced during
induction.

Machine learning has also been used in several other areas of
discourse analysis.  For example, learning has been used to develop
rules for structuring discourse into multi-utterance segments.  
\shortciteA{grosz&hirschberg92} use the classification and
regression tree system {\sc cart}~\cite{brieman84} 
to construct decision trees for
classifying aspects of discourse structure from intonational feature
values.  \shortciteA{LitmanPassonneau95} and \shortciteA{PassonneauLitman97} use the system C4.5 to
construct decision trees for classifying utterances as discourse
segment boundaries, using features relating to prosody, referential
noun phrases, and cue phrases.  In addition, C4.5 has been used to
develop anaphora resolution algorithms, by training on corpora tagged
with appropriate discourse information~\cite{Aone95}.  Similarly,
\shortciteA{McCarthy95} use C4.5 to learn decision
trees to classify pairs of phrases as coreferent or not.  
\shortciteA{SoderlandLehnert94} use the machine learning program
ID3 (a predecessor of C4.5) to support corpus-driven knowledge
acquisition in information extraction.  Machine
learning often results in algorithms that outperform manually derived
alternatives~\cite{LitmanPassonneau95,PassonneauLitman97,Aone95,McCarthy95}, although 
statistical inference is not always used to evaluate the significance of the performance
differences.

Finally, machine learning has also been used with great success in
many other areas of natural language processing.
As discussed above, the work of most researchers in discourse analysis
has concentrated on the direct application of existing symbolic
learning approaches (e.g., C4.5), and on the comparison of learning
and manual methods.  While researchers in other areas of natural
language processing have also addressed these issues, they have in
addition applied a much wider variety of learning approaches, and have
been concerned with the development of learning methods
particularly designed for language processing.  A recent survey of 
learning for natural language~\cite{Wermter96}
illustrates both the type of learning
approaches that have been used and modified
(in particular, symbolic, connectionist, statistical, and
hybrid approaches), as well as the scope of the 
problems that have proved amenable to the use of learning techniques
(e.g., grammatical inference, syntactic disambiguation, and word sense
disambiguation).
\section{Conclusion}
\label{conclusion}

This paper has demonstrated the utility of machine learning techniques
for cue phrase classification.  Machine learning supports the
automatic generation of linguistically viable classification models.
When compared to manually derived models already in the literature,
many of the learned models contain new linguistic insights and perform
with at least as high (if not higher) accuracy.  In addition, the
ability to automatically construct classification models makes it
easier to comparatively analyze the utility of alternative feature
representations of the data.  Finally, the ease of retraining makes
the learning approach more scalable and extensible than manual
methods.

A first set of experiments were presented that used the machine
learning programs {\sc cgrendel}~\cite{CohenIMLC92,CohenIJCAI93} and
C4.5~\cite{Quinlan93} to induce classification models from the
preclassified cue phrases and their features that were used as
training data by \shortciteA{hl93}.  These results were then evaluated with
the same testing data and methodology used by \shortciteA{hl93}.  A second
group of experiments used the method of cross-validation to both train
and test from the testing data used by \shortciteA{hl93}.  A third set of
experiments induced classification models using the new feature {\it
token}. A fourth set of
experiments induced classification models using the new classification
{\it unknown}.

The experimental results indicate that several learned classification
models (including extremely simple one feature models) have
significantly lower error rates than the models developed by \shortciteA{hl93}.  One
possible explanation is that the hand-built classification models were
derived using very small training sets; as new data became available,
this data was used for testing but not for updating the original
models.  In contrast, machine learning in conjunction with
cross-validation (Experiment Set 2) supported the building of
classification models using a much larger amount of the data for
training.  Even when the learned models were derived using the same
small training set (Experiment Set 1), the results showed that the
learning approach helped guard against overfitting on the training
data.  

While the prosodic classification model developed by Hirschberg and Litman
demonstrated the utility of combining phrasal position with phrasal
composition and accent, the best performing prosodic models of
Experiment Sets 1 and 2 demonstrated that phrasal position was in
fact even more useful for predicting cue phrases when used by itself.
The other high performing classification models of Experiment Set 2
also demonstrated the utility of classifying cue phrases based on the
prosodic feature {\it length} and the textual feature {\it preceding
cue phrase}, in combination with other features.  

Just as the machine learning approach made it easy to retrain when new
training examples became available (Experiment Set 2), machine
learning also made it easy to retrain when new features become
available.  In particular, when the value of the feature {\it token}
was added to all the representations in Experiment Set 2, it was
trivial to relearn all of the models (Experiment Set 3).  Allowing
the learning programs to treat cue phrases individually further
improved the accuracy of the learned classification models, and added
to the body of linguistic knowledge regarding cue phrases.
Experiment Set 3 demonstrated that while not useful
by themselves for classifying all cue phrases, 
the prosodic features based on phrasal length, phrasal composition, and
accent, and textual features based on adjacent cue phrases,
succeeding position, and part-of-speech, were in fact useful 
when used only in conjunction with the feature {\it token}.

A final advantage of the machine learning approach is that the ease of
inducing classification models from many different sets of features
supports an exploration of the comparative utility of different
knowledge sources.  This is especially useful for understanding the
tradeoffs between the accuracy of a model and the set of features that
are considered.  For example, it might be worth the effort to code a
feature that is not automatically obtainable or that is expensive to
automatically obtain if adding the feature 
results in a significant improvement in performance.

In sum, the results of this paper suggest that machine learning is a
useful tool for cue phrase classification, when the amount of data
precludes effective human analysis, when the flexibility afforded by
easy retraining is needed (e.g., due to additional training examples,
new features, new classifications), and/or when an analysis goal is to
gain a better understanding of the different aspects of the data.

Several areas for future work remain.  First, there is still room for
performance improvement.  The error rates of the best performing
learned models, even though they outperform the manually derived
models, perform with error rates in the teens.  Note that only the
features that were coded or discussed by \shortciteA{hl93} were considered
in this paper. It may be possible to further lower the error rates by
considering new types of prosodic and textual features (e.g., other contextual
textual features~\cite{Siegel93}, or features that have
been proposed in connection with the more general topic of discourse
structure), and/or by using different kinds of learning methods.
Second, Experiment Set 4 (and the previous literature) show that as
yet, there are no models for predicting when a cue phrase usage should
be classified as {\it unknown}, rather than as {\it discourse} or {\it
sentential}.  Again, it may be possible to improve the performance of
the existing learned models by considering new features and/or
learning methods, or perhaps performance could be improved by
providing more training data.  Finally, it is currently an open
question whether the textual models developed here, which were based
on transcripts of speech, are applicable to written texts.  Textual
models thus need to be developed using written texts as training data.
Machine learning should continue to be a useful tool for helping to
address these issues.

\appendix
\label{app}
\section{C4.5 Results for Experiment Sets 2 and 3}

Tables~\ref{prosodic-c-c},~\ref{textual-c-c} and~\ref{combined-c-c}
present the C4.5 error rates for Experiment Sets 2 and 3.  The C4.5
results for Experiment Set 2 are shown in the ``Non-Tokenized''
columns.  A comparison of Tables~\ref{prosodic-c-c} and~\ref{prosodic}
shows that except for A in the larger test set, the C4.5 prosodic
error rates fall within the {\sc cgrendel} confidence intervals. A
similar comparison of Tables~\ref{textual-c-c} and~\ref{textual} shows
that except for O-P in the larger test set, the C4.5 textual error
rates fall within the {\sc cgrendel} confidence intervals.  Finally, a
comparison of Tables~\ref{combined-c-c} and~\ref{combined} shows that
the C4.5 error rate of {\it speech-text} falls within the {\sc
cgrendel} confidence interval.  The fact that comparable {\sc
cgrendel} and C4.5 results are generally obtained suggests that the
ability to automate as well as to improve upon manual performance is
not due to the specifics of either learning program.

\begin{table*}[tb]
{\scriptsize
\begin{center}
\begin{tabular}{|l || r | r || r| r |} \hline\hline
Model	&\multicolumn{2}{ c||} {Classifiable Cue Phrases (N=878)} &\multicolumn{2}{c|} {Classifiable Non-Conjuncts (N=495)} \\ \hline\hline 
		        &Non-Tokenized	&Tokenized (+) &Non-Tokenized 	&Tokenized (+) \\ \hline\hline 
P-L			&32.5	&31.7	&32.2	&31.4	 \\ \hline 
P-P			&16.2	&18.4	&18.8	&19.0	 \\ \hline 
I-L			&25.6	&26.8	&25.6	&25.6	 \\ \hline 	
I-P			&25.9	&26.3 	&19.4	&18.8	 \\ \hline 
I-C			&36.5	&36.6 	&35.8	&32.8	 \\ \hline 
A			&40.7	&40.7	&29.6	&29.2	 \\ \hline 
A*			&28.3	&26.7	&28.8	&31.2	 \\ \hline  \hline
prosody			&16.0	&15.2	&19.4	&16.0	 \\ \hline  
hl93features		&30.2	&29.0	&18.8	&18.8	 \\ \hline 
phrasing 		&15.9	&15.2	&18.0	&17.4	 \\ \hline  
length			&24.8	&24.4	&26.2	&24.2	 \\ \hline  
position		&18.1	&18.0	&19.6	&17.6	 \\ \hline  
intonational		&16.8	&16.6	&18.8	&19.8	 \\ \hline 
intermediate		&21.2	&22.3	&21.6	&18.4	 \\ \hline  
\end{tabular}
\caption{\label{prosodic-c-c} Error rates (\%) of the 
C4.5 prosodic classification models, testing data. (Training and testing were done from the multiple cue phrase corpus using cross-validation.) }
\end{center}
}
\end{table*}

\begin{table*}[tb]
{\scriptsize
\begin{center}
\begin{tabular}{|l || r | r || r| r |} \hline\hline
Model	&\multicolumn{2}{ c||} {Classifiable Cue Phrases (N=878)} &\multicolumn{2}{c|} {Classifiable Non-Conjuncts (N=495)} \\ \hline  \hline 
		        &Non-Tokenized	&Tokenized (+) &Non-Tokenized 	&Tokenized (+) \\ \hline\hline
C-P			&40.7	&39.3	&39.2	&33.6	\\ \hline	 
C-S			&40.7	&39.9	&39.2	&39.2	\\ \hline
O-P			&40.7	&35.7	&18.6	&14.6	\\ \hline
O-P*			&18.4	&20.3	&17.2	&15.0	\\ \hline
O-S			&35.0	&31.6	&31.8	&31.8	\\ \hline
O-S*			&34.4	&32.5	&31.0	&32.4	\\ \hline
POS			&40.7	&34.7	&41.8	&31.8	\\ \hline\hline
text			&19.0	&20.6	&20.0	&15.0	\\ \hline
adjacency		&40.9	&39.4	&40.6	&43.6	\\ \hline
orthography		&18.9	&19.3	&17.8	&18.0	\\ \hline
preceding 		&18.7	&19.3	&19.2	&16.0	\\ \hline 
succeeding		&34.1	&32.9	&30.0	&31.8	\\ \hline 
\end{tabular}
\caption{\label{textual-c-c} Error rates (\%) of the 
C4.5 textual classification models, testing data. (Training and testing were done from the multiple cue phrase corpus using cross-validation.) }
\end{center}
}
\end{table*}

\begin{table*}[tb]
{\scriptsize
\begin{center}
\begin{tabular}{|l || r | r || r| r |} \hline\hline
Model	&\multicolumn{2}{ c||} {Classifiable Cue Phrases (N=878)} &\multicolumn{2}{c|} {Classifiable Non-Conjuncts (N=495)} \\ \hline\hline 
		        &Non-Tokenized	&Tokenized (+) &Non-Tokenized 	&Tokenized (+) \\ \hline\hline 
speech-text		&15.3	&13.6	&16.8  		&17.6	\\ \hline 
\end{tabular}
\caption{\label{combined-c-c} Error rates (\%) of the 
C4.5 prosodic/textual classification model, testing data. (Training and testing were done from the multiple cue phrase corpus using cross-validation.) }
\end{center}
}
\end{table*}

The C4.5 results for Experiment Set 3 are shown in the ``Tokenized''
columns of Tables~\ref{prosodic-c-c},~\ref{textual-c-c}
and~\ref{combined-c-c}. Comparison with
Tables~\ref{prosodic+},~\ref{textual+} and~\ref{combined+} shows that
the error rates of C4.5 and {\sc cgrendel} are not as similar as 
in Experiment Set 2.  However, the error rates
reported in the tables use the default C4.5 and {\sc cgrendel} options
when running the learning programs. Comparable performance between the
two learning programs can in fact generally be achieved by overriding
one of the default C4.5 options.  As detailed by \shortciteA{Quinlan93}, the
default C4.5 approach -- which creates a separate subtree for each
possible feature value -- might not be appropriate when there are many
values for a feature.  This situation characterizes the feature {\it
token}.  When the C4.5 default option is changed to allow feature
values to be grouped into one branch of the decision tree, 
the problematic C4.5 error rates do indeed improve.
For example, the A+ error rate for the classifiable non-conjuncts
changes from 29.2\% (Table~\ref{prosodic-c-c}) to 11\%, which is
within the 12.8\% $\pm$ 3.1\% {\sc cgrendel} confidence interval
(Table~\ref{prosodic+}).

\acks{I would like to thank William Cohen and Jason Catlett for their helpful
comments regarding the use of {\sc cgrendel} and C4.5, and
Sandra Carberry, Rebecca Passonneau, and the three anonymous JAIR reviewers
for their helpful comments on this paper.
I would also like to thank William
Cohen, Ido Dagan, Julia Hirschberg, and Eric Siegel for comments on a
preliminary version of this paper~\cite{Litman94}.}

\vskip 0.2in

\end{document}